\documentclass[journal]{IEEEtran}

\usepackage[colorlinks,urlcolor=blue,linkcolor=blue,citecolor=blue]{hyperref}
\usepackage{color,array}
\usepackage{times}
\usepackage{epsfig}
\usepackage{graphicx}
\usepackage{amsmath}
\usepackage{amssymb}
\usepackage{array}
\usepackage{xspace}
\usepackage{subcaption}
\usepackage{paralist}
\usepackage{url}
\usepackage{multirow}
\usepackage{booktabs}
\usepackage[norule,symbol,perpage]{footmisc}
\usepackage{comment}
\usepackage{array}
\newcolumntype{P}[1]{>{\centering\arraybackslash}p{#1}}
\usepackage[margin=0.5in]{geometry}
\usepackage{textcomp}
\usepackage{pgfplots}
\pgfplotsset{width=10cm,compat=1.9}
\usepackage{cellspace}
\begin{document}

\title{Learning Domain and Pose Invariance for Thermal-to-Visible Face Recognition}

\author{Cedric Nimpa Fondje$^1$,  
\and
Shuowen Hu$^2$, 
\and
Benjamin S. Riggan$^1$\\
\and
\vspace{5mm}
$^1$University of Nebraska-Lincoln, 1400 R St, Lincoln, NE 68588\\
$^2$CCDC Army Research Laboratory, 2800 Powder Mill Rd., Adelphi, MD 20783\\
\and
\vspace{2mm}
\emph{Corresponding authors: cedricnimpa@huskers.unl.edu, briggan2@unl.edu}
}

\maketitle

\begin{abstract}
Interest in thermal to visible face recognition has grown significantly over the last decade due to advancements in thermal infrared cameras and analytics beyond the visible spectrum.
Despite large discrepancies between thermal and visible spectra, existing approaches bridge domain gaps by either synthesizing visible faces from thermal faces or by learning the cross-spectrum image representations. These approaches typically work well with frontal facial imagery collected at varying ranges and expressions, but exhibit significantly reduced performance when matching thermal faces with varying poses to frontal visible faces. We propose a novel Domain and Pose Invariant Framework that simultaneously learns domain and pose invariant representations. Our proposed framework is composed of modified networks for extracting the most correlated intermediate representations from off-pose thermal and frontal visible face imagery, a sub-network to jointly bridge domain and pose gaps, and a joint-loss function comprised of cross-spectrum and pose-correction losses. 
We demonstrate efficacy and advantages of the proposed method by evaluating on three thermal-visible datasets: ARL Visible-to-Thermal Face, ARL Multimodal Face, and Tufts Face. 
Although DPIF focuses on learning to match off-pose thermal to frontal visible faces, we also show that DPIF enhances performance when matching frontal thermal face images to frontal visible face images.

\end{abstract}

\begin{IEEEkeywords}
heterogeneous face recognition, thermal-to-visible face recognition, biometrics, thermal imaging.
\end{IEEEkeywords}

\section{Introduction}

\IEEEPARstart{F}{ace} Recognition (FR) is perhaps the most widely used biometric modality due to its ease of acquisition and noninvasive nature in conjunction with its high accuracy.
%among the most widely used and discriminative biometric modalities---face,fingerprint, and iris%.
Although FR systems still require a minimum interocular distance (IOD), e.g., ideally greater than 60 pixels between the eyes~\cite{6512016}, using faces for identification has the advantage of better standoff acquisition compared to iris and fingerprint modalities.
In fact, many consumer electronics, social media platforms, and military or law enforcement surveillance tools have incorporated either face identification and/or face verification capabilities~\cite{ABISreport}.

\begin{figure}
%\centering
\includegraphics[width=0.50\textwidth]{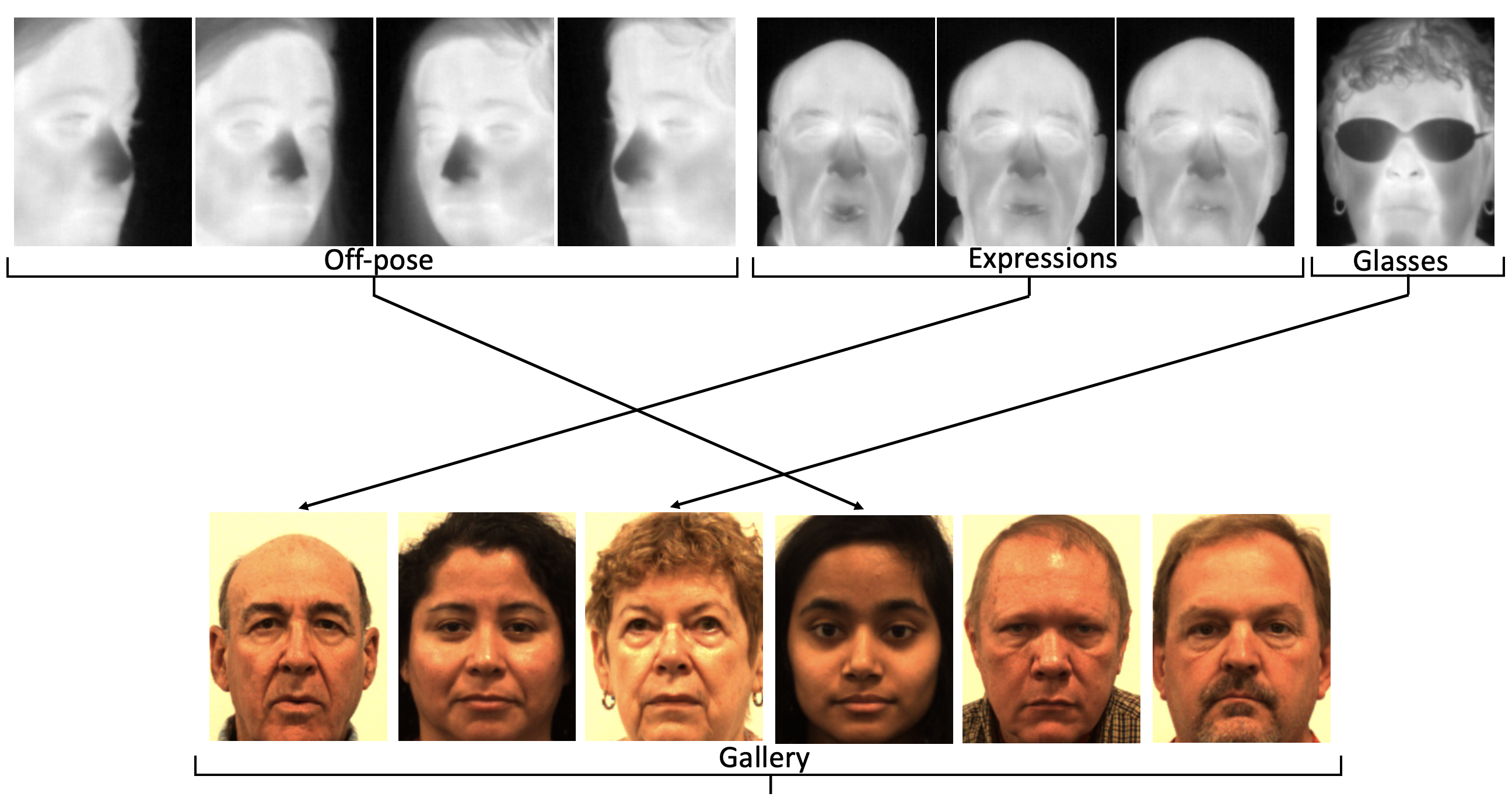}
\caption{Thermal-to-visible face matching typical objective. Thermal probe imagery and visible gallery imagery are both sampled from ARL-VTF dataset.}
\label{fig:verificatiion}
\end{figure}

FR has shown significant improvement over the recent decades due to the availability of large-scale annotated face datasets (e.g. MS-Celeb-1M~\cite{Guo2016MSCeleb1MCO}, LFW~\cite{LFWTech}, MegaFace Challenge 1 and 2~\cite{KemelmacherShlizerman2016TheMB, 8099846}, ARL-VTF~\cite{Poster_2021_WACV}) and the growth of modern deep learning models~\cite{10.5555/2969033.2969049, 6909616}. However, all deployed FR systems perform matching in the visible spectrum, and emerging cross-spectrum FR capabilities have not yet reached sufficient accuracy, especially under variable poses, to be deployed for operational use.

Despite recent advances toward perspective invariance, large pose variations between probe and gallery face images still result in decreased face recognition performance. 
%apparent differences in perspective (or pose) result in severe reductions in performance. 
Sengupta et al.~\cite{7477558} demonstrated that most FR models exhibit a drop in performance of at least 10\% in profile-to-frontal face matching compared to frontal-to-frontal face matching in the visible spectrum. This drop becomes more significant when matching faces across different spectra. To address this problem, scientists and engineers have investigated models, including domain adaptation methods, for heterogeneous facial recognition (HFR) to match non-frontal and frontal face imagery from divergent spectra (domains): near infrared (NIR) to visible~\cite{10.1007/978-3-540-74549-5_55, 5597000, 6595898} or thermal infrared to visible~\cite{Hu:15, Gurton:14, Short:15}. 

In this paper, we primarily focus on thermal-to-visible FR by presenting a novel state-of-the-art domain adaptation framework that is simultaneously robust to both spectra and pose. The design hypothesis of our proposed framework is based on the assumption that there exists a \emph{pose invariant} mapping between the visible and the thermal image representations. 
%and the framework aims to learn a such mapping. 
Although Residual Spectral Transform (RST)~\cite{9304937} demonstrated great performance in thermal-to-visible FR by achieving 96\% in rank-1 ID when using frontal face imagery with varying expressions, its performance significantly dropped to 29.91\% when using off-pose face images.
Our new framework transforms image representations extracted from different spectra using a CNN-based sub-network---DPIT (Domain and Pose Invariance Transform)---to map the disparate representations to a common latent embedding subspace.  Our framework significantly outperforms \cite{9304937} by optimizing base architecture, sub-network transform, and loss functions to directly and jointly address both domain and pose gaps. Despite our primary aim to address matching non-frontal thermal face imagery (probe samples) with frontal visible imagery (gallery samples), we also indirectly enhance frontal thermal-to-visible FR.

Recently, there have been two types of techniques for thermal-to-visible FR: (1) learning domain invariant image representations in feature or latent space and (2) image-to-image translation. Invariant representation models typically extract image representations from each of the two domains first, and then attempt to learn non-linear mapping(s) that yield sufficiently discriminative embedding representations across disparate domains. Such models were recently implemented using different approaches such as using partial least squares (PLS) in a one-vs-all setting~\cite{Hu:15}, using regression neural networks between the thermal representation and the corresponding visible representation~\cite{DBLP:conf/bmvc/SarfrazS15}, or using coupled auto-associative neural networks (CpNN)~\cite{7270978}. 

Alternatively, image-to-image translation models aim to synthesize visible faces from the corresponding thermal faces (or vice versa), then extract embedding representations from  both synthesized visible probe imagery and visible gallery imagery mainly using existing face feature extraction/matching models such as VGG Face~\cite{vggface2015}. Some of these models include: Pix2Pix~\cite{pix2pix2017} that uses a U-net architecture when synthesizing visible, GANVFS~\cite{8272687} which trains a synthesis network using identity and perceptual losses, SAGAN~\cite{8987329} which is an adaptation of self-attention~\cite{2d4b2d81abb74b56938c71cd38e10d25} to CycleGAN~\cite{8237506}.

In our proposed framework, we learn invariant representations as this is typically less complex to implement (e.g., no discriminator networks and no decoding network) than image-to-image translation. When comparing the proposed invariant representation method with other recent image-to-image translation methods, we observe that the proposed method achieves better discriminability overall.

The proposed domain adaptation framework is composed of a feature extraction block (or base architecture), a sub-network that learns the mapping between the extracted thermal features and the visible features, and a combination of new cross-spectrum and pose-correction losses to guide the features during the identification process. Our approach effectively performs face frontalization performed in the embedding space rather than directly frontalizing images. Moreover, our proposed approach extends beyond a naive combination of frontalization and cross-spectrum transformation by seamlessly addressing both domain and pose gaps at the same time.

In summary, we propose a novel domain adaptation framework for thermal-to-visible FR that is robust to pose variations through the following contributions:
\begin{compactitem}
    \item modified base architectures---VGG16 \cite{Simonyan15} and Resnet50 \cite{He2016IdentityMI}---to extract the most correlated and invariant representations possible across thermal and visible domain, 
    \item a new feature mapping sub-network, DPIT, to further bridge domain and pose gaps between the extracted image representations,
    \item a new joint loss function---a linear combination of cross-spectrum and the pose-correction losses---designed (1) to preserve identity across thermal and visible domain and (2) to encourage pose invariant in the embedding space,
    \item extensive analysis using ARL-VTF~\cite{Poster_2021_WACV}, ARL-MMF Face~\cite{article}, and Tufts Face~\cite{8554155} datasets to compare the proposed approach against recent state-of-the-art approaches, 
    \item ablation studies for understanding the effects of embedding size and pose-correction loss on face verification performance.
\end{compactitem}

When comparing our proposed model to other state-of-the-art invariant representations methods~\cite{cao2018Dream, 9304937}, and image-to-image translation methods~\cite{9358101, 7477447, DBLP:journals/ijcv/SarfrazS17, XingHFF}, our proposed framework achieves new state-of-the-art results when matching profile thermal face imagery to frontal visible face imagery on ARL-VTF, ARL-MMF, and Tufts Face datasets.

The remainder of the paper is structured as follows: In Section~\ref{sec:related}, we review some related works on visible to thermal face recognition. Section~\ref{sec:method} presents details of our proposed model. The different datasets used as well as some corresponding protocols are described in Section~\ref{sec:dataset}. Extensive analysis/experiments and results are presented in Section~\ref{sec:experiment}. Finally, we conclude this paper with a brief summary in section~\ref{sec:conclusion}.

\section{Related Work}
\label{sec:related}
In this section, we present and summarize prior work related to face frontalization and thermal-to-visible face recognition.

\subsection{Face Frontalization for Face Recognition}
Both within-spectrum and cross-spectrum FR models suffer from large pose variations. To boost performance in profile-to-frontal face matching, Masi et al.~\cite{MTHLM:2016:dowe} suggested training existing CNN-based models with datasets that have more faces with large pose variations and that differ in 3D shape. Some more recent studies have suggested face frontalization~\cite{Hassner2015EffectiveFF,8578333,7410798,XingHFF, cao2018Dream} to mitigate effects of pose variations on FR performance, rather than augmenting the train and test dataset. Hassner et al.~\cite{Hassner2015EffectiveFF} proposed a single 3D unmodified model to frontalize profile input face images. Although this approach enhanced visible FR and gender estimation performance by approximately 4\%, this is dependent on precise and accurate localization of 48 local descriptors around eyes, nose, and mouth. Furthermore, Yin et al.~\cite{towards-large-pose-face-frontalization-in-the-wild} proposed the FF-GAN model which incorporates the 3DMM (3D Morphable Face Model) into a generative model to provide the desired shape and a better appearance. By doing this, the FF-GAN model learns faster while preserving the face texture and shape. The proposed framework differs from these models by addressing a more challenging problem involving different spectra and uses image representations rather than synthesized images to jointly mitigate pose and domain gaps.
Di et al.~\cite{XingHFF} proposed a cross-spectrum face frontalization model using domain agnostic learning along with a dual-path encoder/decoder architecture to improve discrimination across the thermal and the visible domain. Similar to~\cite{XingHFF}, we proposed a cross-spectrum face frontalization framework, but instead of synthesizing frontal visible face imagery from off-pose (e.g., profile) thermal images, our model directly learns image representations that are robust to pose and domain variations.
Sagonas et al.~\cite{7410798} proposed a model to simultaneously localize landmarks from off-pose faces to iteratively reconstruct a frontal view. Unlike this approach, our proposed framework makes no use of landmark and pose information. Instead, we use extracted image representations to learn a discriminative mapping between off-pose and frontal faces.

\subsection{Pose-Robust Feature Learning in Face Recognition} 
Some previous work include Cao et al.~\cite{cao2018Dream} who suggested frontalizing visible profile face imagery in the latent embedding space using the deep residual equivariant mapping (DREAM) to bridge the gap between frontal faces and corresponding profile faces using the pose information as a regularizer. On the contrary to DREAM that leverages head pose information by using a pose estimation sub-network, we propose a pose-correction loss function to increase the similarity between frontal and profile image representations \emph{without any explicit dependence on pose information, fiducial landmarks, or 3D models}. Moreover, our proposed framework jointly optimizes our cross-spectrum and pose-correction losses to simultaneously reduce both domain and pose variations.
Other pose-invariant feature learning methods such as DFN (Deformable Face Net)~\cite{8756575} exploits deformable convolutional modules to align feature embedding extracted from imagery exhibiting variable pose. During training, DFN uses paired images (from two different domains) of same subject and minimize intra-class variations.

\subsection{Feature-based Thermal-to-Visible Face Recognition}
Conventional thermal-to-visible face recognition algorithms exploit image representations extracted from the thermal and visible face imagery to learn a mapping that brings these representations into a common latent subspace.
Choi et al.~\cite{10.1117/12.920330} proposed an approach based on a discriminant analysis with partial least square (PLS) to match the thermal face imagery to the thermal to the corresponding visible face imagery signature. This approach was improved in~\cite{Hu:15} where Hu et al. illustrated that using a one-vs-all framework with partial least square (PLS) classifiers provided more distinction when matching thermal and visible HOG features.
Sarfraz et al.~\cite{Sarfraz2015DeepPM} proposed a deep perceptual mapping (DPM) model to learn the thermal features from the extracted visible features. This DPM model was basically a trainable neural network that regresses the thermal features (e.g., HOG or DSIFT) from the visible features. The authors also showed that performing a dimensionality reduction (e.g., PCA) after extracting the thermal DSIFT features using a three-layer DPM was very effective in estimating the corresponding visible features. However, this approach does not work well for matching off-pose face images with frontal face images since HOG and DSIFT features are not pose invariant.
In~\cite{7270978}, coupled neural networks were proposed to learn how to extract the closest similar latent features between the visible and the polarimetric thermal face imagery by minimizing the mean square error between the polarimetric thermal and the visible features. 
The feature extraction network was similar to the VGG architecture used in~\cite{8411218} and included some global average pooling to prevent over-fitting. While global pooling may potentially reduce the gaps between off-pose thermal and frontal visible image representations, it also removes a significant amount of contextual information relating to identity. Therefore, we opt to alleviate potential over-fitting in alternative ways, such as guiding the architecture to ignore discriminative high frequency visible textures that are not present in thermal imagery.
More recently, Nimpa Fondje et al.~\cite{9304937} introduced a CNN-based framework that also leverages deep features across the thermal and the visible modalities. With this framework, features were extracted from truncated versions of pretrained VGG16 and Resnet50 networks. Then, to bridge the domain gap between the extracted visible and thermal representations the authors proposed the residual spectral transform (RST) along with domain invariance and cross-domain loss functions. 
Despite outperforming state-of-the-art methods for thermal-to-visible FR on frontal imagery, like~\cite{7270978,Sarfraz2015DeepPM, 5597000, Short:15}, RST still lacks discriminability with large pose variations between probe and gallery face images; the rank-1 identification rate for RST drops by more than 60\%.

\subsection{Synthesis-based Thermal-to-Visible Face Recognition}
Synthesis-based approaches have also been effective for thermal to visible face recognition. These methods aims to estimate the appearances of individual faces as if acquired from a visible camera by synthesizing a visible-like face image from the thermal face image input, and then matching the synthesized image against the visible spectrum gallery face images.
In~\cite{8272687} Zhang et al. presented an encoder-decoder style generator within a GAN-based approach to synthesize visible face imagery from corresponding thermal imagery. This particular GAN optimized a combination of adversarial, perceptual, and identity loss functions to mitigate the domain gap. Riggan et al.~\cite{8354114} proposed a synthesis method that used the local and global facial regions to improve the quality of synthesized visible face imagery. Another synthesis model was presented by Di et al.~\cite{Di2018PolarimetricTT} that combines the extracted features from thermal-to-visible and visible-to-thermal synthesized images for face verification. He et al.~\cite{8941241} recently proposed a synthesis model that used texture impainting as well as pose-correction. More recently Di et al.~\cite{9358101} proposed to use the facial attributes to guide the synthesis of a frontal visible face from the corresponding frontal thermal faces.
Zhao et al.~ proposed the Pose Invariant Model (PIM)~\cite{8578333}, to synthesize frontal faces from their corresponding profile faces through a dual pathway generative adversarial network (GAN) that simultaneously learned both global structure and local representations of face imagery using unsupervised adversarial optimization. Subsequently, Cao et al.~\cite{Cao2018LearningAH} improved upon PIM~\cite{8578333} by proposing High Fidelity PIM (HF-PIM) in which they introduced a new texture warping strategy while leveraging dense correspondence fields between 2D and 3D spaces. Their approach not only preserves subject identity, but also improves texture of synthesized frontal faces images making them more photo realistic.

Both the feature-based and synthesis-based method we reviewed above provide very good results on dataset with frontal and variable expression thermal imagery, but there is a significant drop in performance when attempting to use same model on a dataset with profile face thermal imagery. To this end, our proposed method not only show robustness to pose variations for thermal-to-visible FR, but also achieves state-of-the art performance for frontal thermal-to-visible FR.

\begin{table*}[ht]
    \begingroup
    \caption{Reference notations for base architectures, where the number (e.g., 1) indicates the block number and the letter (e.g., a) indicates the group of convolutional operators within each block.  Each block is given as a group of convolution and/or pooling operators with filters of size $H \times W$ followed by `$ c_{out}~/~s$' indicating number of filters and stride, respectively. The rightmost`$\times~p$' notation specifies the number of convolutional groups per block.}
    \label{tab:feat_extract}
    \endgroup
    \begin{subtable}[t]{0.45\textwidth}
        \centering
        \begin{tabular}[t]{|P{1cm}|Sc|}
        \hline
        \textbf{Blocks} & \textbf{Operators}\\
        \hline
        1a & $\begin{bmatrix} 7\times7~,~64~/~2\\ 3\times3~Pool\\ \end{bmatrix}\times1$\\
        \hline
        2a--c & $\begin{bmatrix} 1\times1~,~64~/~1\\ 3\times3~,~64~/~1\\ 1\times1~,~256~/~1\\ \end{bmatrix}\times3$\\
        \hline
        3a--d & $\begin{bmatrix} 1\times1,~128~/~2\\ 3\times3~,~128~/~1\\ 1\times1~,~512~/~1\\ \end{bmatrix}\times4$\\
        \hline
        4a--f & $\begin{bmatrix} 1\times1~,~256~/~2\\ 3\times3~,~256~/~1\\ 1\times1~,~1024~/~1\\ \end{bmatrix}\times6$\\
        \hline
        5a--c & $\begin{bmatrix} 1\times1~,~512~/~2\\ 3\times3~,~512~/~1\\ 1\times1~,~2048~/~1\\ \end{bmatrix}\times3$\\
        \hline
       \end{tabular}
       \caption{ResNet50}
       \label{tab:base_resnet50}
    \end{subtable}
    \hfill
    \begin{subtable}[t]{0.45\textwidth}
        \centering
        \begin{tabular}[t]{|P{1cm}|Sc|}
        \hline
        \textbf{Blocks} & \textbf{Operators}\\
        \hline
        1 & $\begin{bmatrix} 3\times3~,~64~/~1 \\ 3\times3~,~64~/~1\\ 2\times2~Pool\\ \end{bmatrix}\times1$\\
        \hline
        2 & $\begin{bmatrix} 3\times3~,~128~/~2\\ 3\times3~,~128~/~2\\ 2\times2~Pool\\ \end{bmatrix}\times1$\\
        \hline
        3 & $\begin{bmatrix} 3\times3~,~256~/~2\\ 3\times3~,~256~/~2\\ 3\times3~,~256~/~2\\ 2\times2~Pool\\ \end{bmatrix}\times1$\\
        \hline
        4 & $\begin{bmatrix} 3\times3~,~512~/~2\\ 3\times3~,~512~/~2\\ 3\times3~,~512~/~2\\ 2\times2~Pool\\ \end{bmatrix}\times1$\\
        \hline
        5 & $\begin{bmatrix} 3\times3~,~512~/~2\\ 3\times3~,~512~/~2\\ 3\times3~,~512~/~2\\ 2\times2~Pool\\ \end{bmatrix}\times1$\\
        \hline
        \end{tabular}
        \caption{VGG16}
        %\label{tab:base_vgg16}
     \end{subtable}
     \vspace{-0.7cm}
\end{table*}

\section{Proposed Method}
\label{sec:method}
In this section, we provide a detailed description of the proposed Domain and Pose Invariance Framework. Specifically, we present the modified base architectures, DPIT sub-network that performs cross-spectrum face frontalization, and cross-spectrum and pose correction loss functions. More importantly, we discuss how these components interact with each other to yield our novel DPIF shown in Fig.~\ref{fig:framework}.

\subsection{Modified Base Architectures}
The base architectures used to extract discriminative image representations from off-pose thermal and frontal visible imagery are motivated by both Resnet50~\cite{He2016IdentityMI} and VGG16~\cite{Simonyan15} architectures (see Table.~\ref{tab:feat_extract}), which are widely used for FR applications. Similar to~\cite{9304937}, we observed that the raw features output by complete VGG16 and Resnet50 networks are less transferable across the visible and the thermal spectrum due to being trained on large-scale visible imagery with rich textural information that is not present in thermal imagery. Since the deep layers in these architectures are most sensitive to the high frequency content representing the fine textural facial details, we determine the best intermediate level feature maps to use for cross-spectrum matching. Therefore, by truncating the base architectures such that the receptive fields are maximized and the extracted visible and thermal image representations are most similar, we are able to quickly alleviate any potential over-fitting to discriminative domain-dependent information (e.g., visible textures) that are absent from thermal face imagery.

Using Resnet50, we found that the most effective block for thermal off-pose to visible frontal matching was the fifth convolutional group from the fourth residual block (i.e., `block\_4e'). Similarly, we determined that the features extracted from the fourth convolutional-pooling block (i.e., `block4\_pool') of VGG16 provide better performance in cross-spectrum matching. We used truncated VGG16 and truncated Resnet50 to initially extract common features to match profile thermal face imagery with corresponding frontal visible face imagery. Interestingly, we determine the optimal feature map dimensions ($H \times W \times C$ -- height by width by channels) for matching thermal off-pose probes to visible frontal gallery to be $14\times14\times1024$ and $14\times14\times512$ for Resnet50 and VGG16, respectively (see ablation study in section~\ref{sec:experiment}). 

Given an image $\mathbf{x}$ from either domain (visible or thermal), let
\begin{equation}
\mathbf{x}_m=\phi_{m}(\mathbf{x}),
\label{eq:mba}
\end{equation}
denote the feature maps from our modified based architecture where $m \in \{$VGG16, Resnet50$\}$ with dimensions $H \times W \times C.$

Following the proposed modified base architecture (Eq.~\ref{eq:mba}), a shared compression layer which consists of single $1 \times 1$ convolutional layer is applied to both streams. This compression layer reduces the effect of the noise propagation and also reduces the number of learnable parameters in the subsequent layers by compressing the base architecture feature maps in a channel-wise fashion by a factor of 2 (i.e., a 50\% compression rate). Moreover, through optimizing this shared layer, our model is able to learn a common projection for visible and thermal representations. This compression is given by
\begin{equation}
%\mathcal{F}(\mathbf{x})= \text{Conv}_{c} \circ \text{Conv}_{200} \circ \text{Conv}_{200}(\mathbf{x}) ~, \\
h(\mathbf{x}_m)= tanh(\mathbf{W}_{C/2}' \circledast \mathbf{x}_m + \mathbf{b'}) ~, \\
\label{eq:eq_H}
\end{equation}
where $\circledast$ represents the convolution operator, $\mathbf{W}_{C/2}'$ and $\mathbf{b'}$ are convolution kernel and bias parameters, the subscript $C/2$ denotes the number of output channels, and $tanh(u)=((exp(u) - exp(-u))(exp(u) + exp(-u)))^{-1}$ is the hyperbolic tangent activation function with output range $(-1,~1).$

\subsection{Domain and Pose Invariance Transform (DPIT)}

\begin{figure*}[htb]
\centering
\includegraphics[width=1.0\textwidth]{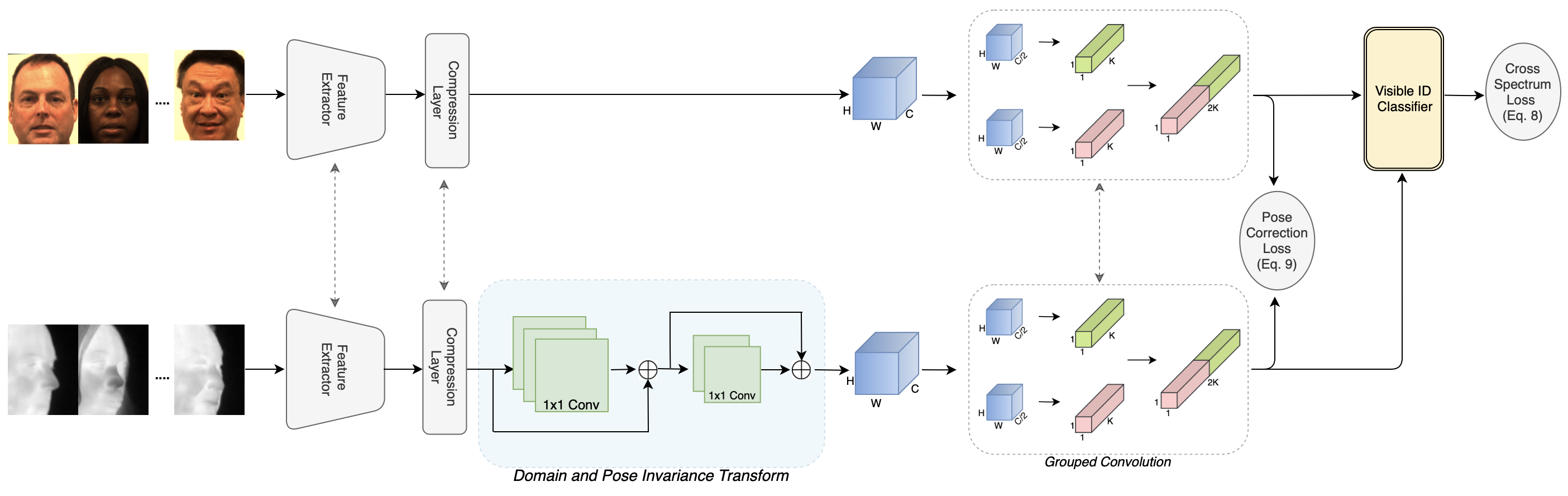}
\caption{Our proposed domain and pose invariant framework (DPIF) which uses the domain and pose invariance transform (DPIT) to transform thermal image representations.  The dashed arrows (gray) indicate shared models parameters.}
\label{fig:framework}
\end{figure*}

The DPIT sub-network---comprising two consecutive residual blocks---simultaneously preserve identity information while effectively performing face frontalization and cross-domain transformation using face embeddings. DPIT transforms the output from the compression layer between thermal and visible representations. These two residual blocks use $1 \times 1$ convolutional layers to map thermal image representations to visible. Then, a grouped convolutional layer provides additional robustness to pose variations.
The first residual block is given by
\begin{equation}
\mathcal{F}(\mathbf{x}_m^t)= f_{c}^{3} \circ f_{200}^{2} \circ f_{200}^{1}(h(\mathbf{x}_m^t)) + h(\mathbf{x}_m^t)~, \\
\label{eq:eq_F}
\end{equation}
with
\begin{equation}
    f_k^\ell(\mathbf{z}) = tanh(\mathbf{W}_k^\ell  \circledast \mathbf{z} + \mathbf{b}^\ell) ~, \\ 
\end{equation}
where the subscript $k$ denotes the number of output units (i.e., channels), $\mathbf{W}_k^\ell$, $\mathbf{b}^\ell$ are convolution kernel and bias parameters for the $\ell^{th}$ layer, and $\mathbf{z}$ represents the input feature maps for layer $\ell$.  We use the superscript $t$ in $x_m^t$ to specify that the feature maps are extracted from thermal imagery (as opposed to super script $v$ for visible imagery).

Similarly, the second residual block is given by
\begin{equation}
\mathcal{G}(\mathbf{x}_m^t)= g_{c}^{2} \circ g_{c}^{1}(\mathcal{F}(\mathbf{x}_m^t)) + \mathcal{F}(\mathbf{x}_m^t)~, 
\label{eq:eq_G}
\end{equation}
with
\begin{equation}
    g_k^\ell(\mathbf{z}) = ReLU(\mathbf{\bar{W}}_k^\ell \circledast \mathbf{z} + \mathbf{\bar{b}}^\ell) ~, \\ 
\end{equation}
where the subscript $k$ denotes the number of output units (i.e., channels), $\mathbf{\bar{W}}_k^\ell$ and $\mathbf{\bar{b}}^\ell$ are convolution kernel and bias parameters for the $\ell^{th}$ layer, and $ReLU(u)=\max(0,u)$ is the rectified linear unit activation function with output range $[0,~1).$

Notice that the two residual blocks in Eqs.~\ref{eq:eq_F} and \ref{eq:eq_G} use two different activation function: $tanh$ and $ReLU$. Although mixing activation functions is not common, we hypothesize the advantage in this context comes from combining the capacity limitations due to $1 \times 1$ convolutional layers, with activation limits imposed by $tanh$, and activation sparsity introduced by $ReLU$.  We considered various combinations of $tanh$ and $ReLU$ activation functions, such as $tanh$ followed by $tanh$, $ReLU$ followed by $ReLU$, and $ReLU$ followed by $tanh$.  However, as shown in Table~\ref{tab:activation_func} we found that $tanh$ followed by $ReLU$ for the two blocks provided the most discriminative performance for thermal-to-visible FR.

Eq.~\ref{eq:eq_G} is defined as the DPIT subnetwork, which is illustrated in Fig.~\ref{fig:framework}.

Following the DPIT transform for thermal representations and following the compressed visible representation according to Eq.~\ref{eq:eq_H}, both visible and thermal representations are mapped using a shared grouped convolution layer:  
\begin{equation}
    \psi(\mathbf{x}_m) = ReLU(\mathbf{V} \circledast_G^{n} \mathcal{T}(\mathbf{x}_m) + \mathbf{c}),
\end{equation}
where $\circledast_G^n$ denotes the grouped convolution with $n$ filter groups,$\mathbf{V}$ and $\mathbf{c}$ denote the weights and biases, and $\mathcal{T}(\mathbf{x}_m)$ represents either $h(x_m^v)$ or $\mathcal{G}(\mathbf{x}_m^t)$ for the visible or thermal streams, respectively.

Since the first two blocks effectively operate on ``patches'' with particular sizes (i.e., receptive fields) associated with each activation, to increase tolerance to perspective variations, we consider a more holistic image representation.  Given the likelihood of over-fitting when matching off-pose thermal faces with frontal visible faces, we use a grouped convolution that uses multiple filter groups to reduce number parameters by a factor related to the number of filter groups and to learn complementary filter groups. Therefore, this increases the computational efficiency and decreases the risk of over-fitting while also providing additional robustness to pose.  

We experimentally determined $n=2$ filter groups to work well within our proposed framework. 

\subsection{Cross-spectrum Loss}
The cross-spectrum loss is used to ensure that the identity of different subjects is preserved during training by simultaneously learning discriminative features in the thermal spectrum for performing the recognition task against visible spectrum gallery images.
\begin{equation}
    \mathcal{L}_{\mathcal{C}} = -\sum \mathbf{y}\cdot\log\hat{\mathbf{y}}(\psi(\mathbf{x}_m^t); \mathbf{\Theta}_v) ~,
\label{eq:cross_spectrum_loss}
\end{equation}
where ${\textbf{y}}$ represents the true identities (labels) and $\hat{\mathbf{y}}(\psi(\mathbf{x}_m^t); \mathbf{\Theta}_v )$ are the predicted identities or labels obtained by feeding the \emph{DPIT~transformed} feature representation $\psi(\mathbf{x}_m^t)$ (i.e., frontalized and `visible-like' image representation) to a classifier~$\hat{\mathbf{y}}(\cdot,\mathbf{\Theta}_{v})$ that is discriminatively trained to perform an identification task using only frontal visible face imagery; mimicking a common constrained enrollment gallery.  Therefore, the proposed cross-spectrum loss is used to optimize and to enhance the DPIT.

\subsection{Pose-Correction Loss}
This loss function aims to push the representations from frontal and off-pose face imagery close to each other by minimizing the $L_2$ distance between them. 

\begin{equation}
    \mathcal{L}_{\mathcal{P}} = \sum\limits_{i} \left\Vert \psi(\mathbf{x}_{m,i}^v) - \psi(\mathbf{x}_{m,i}^t) \right\Vert^2 ~,
\label{eq:pose_loss}
\end{equation}
where $\psi(\mathbf{x}_{m,i}^v)$ corresponds to the holistic image representation extracted from frontal visible face imagery and $\psi(\mathbf{x}_{m,i}^t)$ is the resulting \emph{DPIT~transformed} representations from off-pose thermal face imagery.

Therefore, the proposed joint-loss function which combines Eqs.~\ref{eq:cross_spectrum_loss}-\ref{eq:pose_loss} is given by:
\begin{equation}
    \mathcal{L}= \mathcal{L}_{\mathcal{C}} + \lambda \cdot \mathcal{L}_{\mathcal{P}}~,
    \label{eq:total_loss}
\end{equation}
where $\lambda$ is the loss parameter.

\subsection{Implementation Details}

Most thermal-to-visible face recognition systems require to pre-process the images before using them for training and/or testing. In this work, we first registered and then tightly cropped the thermal and visible images to a $224\times224$ image around the center of the eyes, the base of the nose, and the corners of the mouth. The DPIF was trained in Tensorflow on a single Nvidia GeForce RTX 2080 Ti GPU. 

During training, the base architecture was made not trainable when using either the truncated Resnet50 or VGG16 and the compression was done channel-wise at a 50\% ratio. We first trained the visible ID classifier to be robust on frontal visible faces using the cross entropy loss for 5 epochs. Then we made the Visible classifier not trainable and we trained the thermal stream for 100 epochs using our proposed DPIT sub-network and loss functions. The pose-correction loss was computed during training by simultaneously feeding to the thermal stream and the visible stream the off-pose thermal faces and the corresponding frontal visible faces, respectively. Thus, $\mathcal{L}_{\mathcal{P}}$ was computed from the actual visible feature representation and the predicted visible representation obtained after the grouped convolution block. Similar to~\cite{9304937}, the predicted visible representations and the actual visible representations are matched using the cosine similarity measure during inference. Note that we experimentally found that the optimal value of $\lambda$ was $10^{-5}$ and that an embedding size of 256 was providing the best trade-off between the model performance and the required number of trainable parameters. We show the effect of varying these two parameters in section~\ref{sec:experiment}.

The proposed methodology is not limited to thermal-to-visible face recognition and may also be applied to other spectra, such as NIR (see Appendix~\ref{appendix:casia}).

\section{Datasets}
\label{sec:dataset}
In this section, we provide a brief description of the different datasets/protocols used for analyzing our DPIF for thermal-to-visible FR. More specifically, we describe the ARL-VTF dataset~\cite{Poster_2021_WACV}, the ARL-MMF dataset~\cite{article} and the Tufts Face database~\cite{8554155}, which were used to evaluate our proposed domain adaptation framework.

\subsection{ARL Visible-Thermal Face Dataset}

The ARL-VTF dataset is one of the largest collection of paired conventional thermal and visible images. The ARL-VTF dataset contains over 500,000 images corresponding to 395 unique subjects among which 295 are in the development set (i.e., training and validation sets) and 100 in the test set. In addition to baseline---frontal with neutral expression---imagery, ARL-VTF includes expression, pose (from $-90^\circ$ to $+90^\circ$), and eye-wear (e.g., glasses) variations across multiple subjects.  
Similar to~\cite{9484353}, we evaluated the proposed method using the provided protocols as described below. 

The ARL-VTF dataset includes two gallery sets, denoted as G\_VB0- and G\_VB0+.  Each gallery set is a collection of visible baseline imagery from 100 subjects.  In G\_VB0- none of the subjects were collected wearing glasses.  However, in G\_VB0+, approximately 30 subjects, who typically wear glasses, were collected wearing their glasses and the remaining 70 subjects were collected without any glasses.

The ARL-VTF probe sets are grouped into baseline, expression, pose, and eye-wear conditions.  

\textbf{Baseline:} There are three different probe sets that include baseline thermal imagery subsets from the 100 enrolled subjects: P\_TB0, P\_TB-, and P\_TB+. Here the suffix `0' specifies imagery of subjects who do not have glasses (i.e., 70 subjects),`-' denotes the imagery from subjects who have glass but were not wearing them, and `+' denotes the imagery from subject that have glass and had their glasses on (i.e., 30 subjects). Note that the effect of eye-wear variation in the probe sets is only considered for baseline condition. 

\textbf{Expression:} There are two different probe sets that include thermal imagery with varying expressions: P\_TE0 and P\_TE-.  Similarly, these probe sets are composed of either imagery of subjects who do not have glasses (i.e., 70 subjects) or imagery from subjects who have glass but were not wearing them (i.e., 30 subjects).

\textbf{Pose:} There are two different probe sets that include thermal imagery with varying pose: P\_TP0 and P\_TP-.

\begin{figure*}[htb]
\centering     %%% not \center
\subfloat[Ground truth]{\label{fig:score_a}\includegraphics[width=0.20\textwidth]{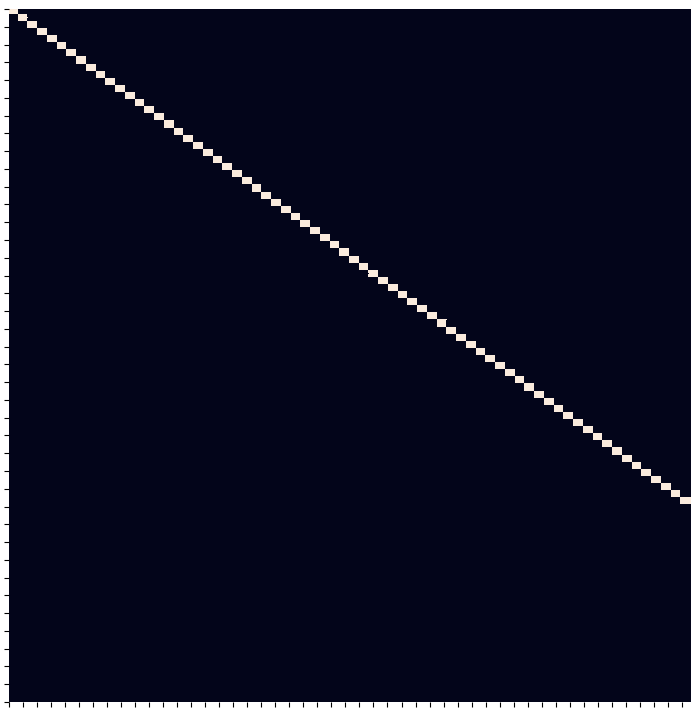}}
\subfloat[Raw]{\label{fig:score_b}\includegraphics[width=0.20\textwidth]{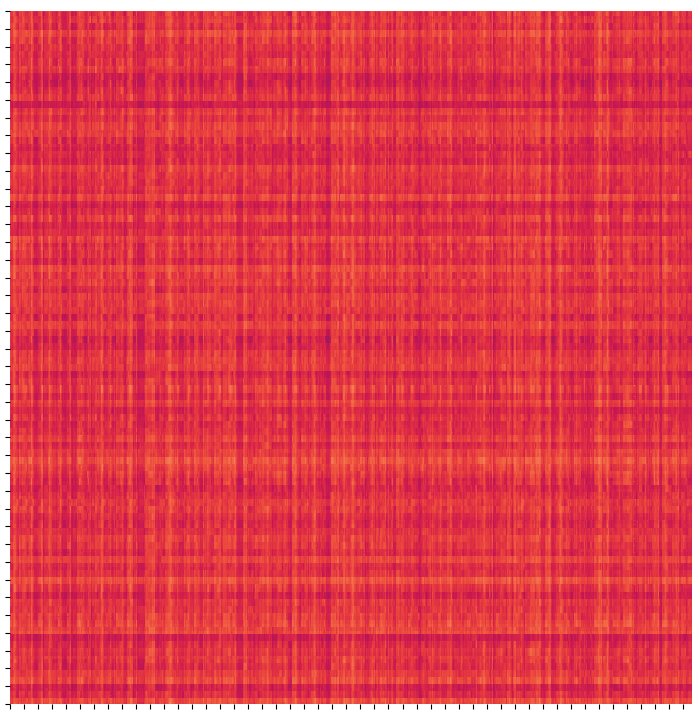}}
\subfloat[Dream~\cite{cao2018Dream}]{\label{fig:score_c}\includegraphics[width=0.20\textwidth]{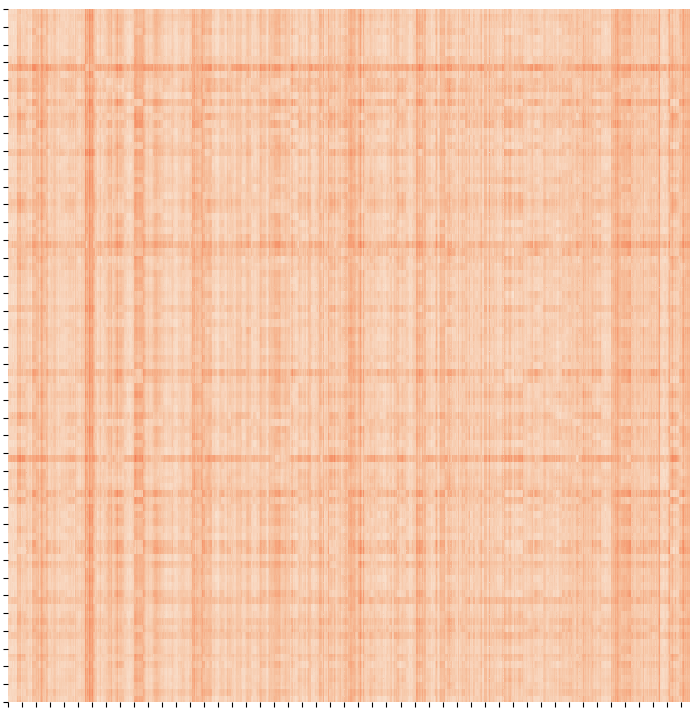}}
\subfloat[RST~\cite{9304937}]{\label{fig:score_d}\includegraphics[width=0.20\textwidth]{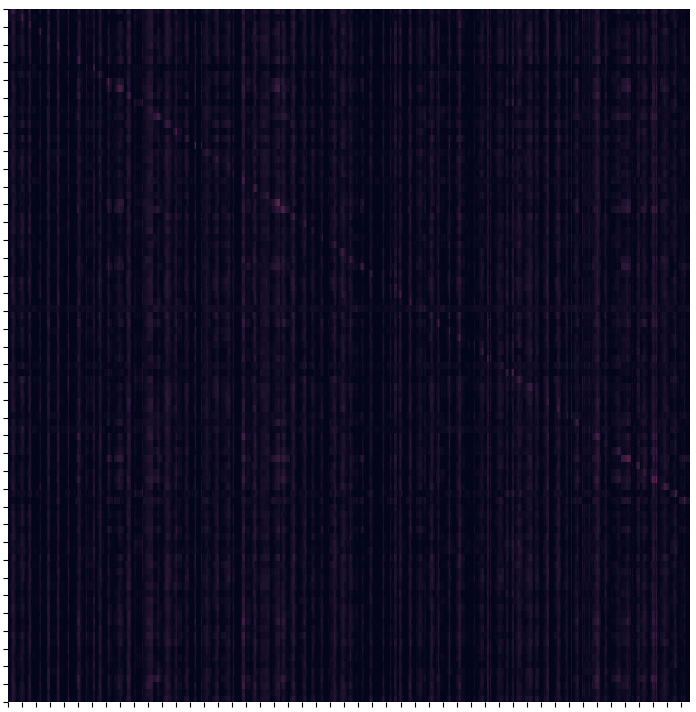}}
\subfloat[DPIT(Ours)]{\label{fig:score_e}\includegraphics[width=0.23\textwidth]{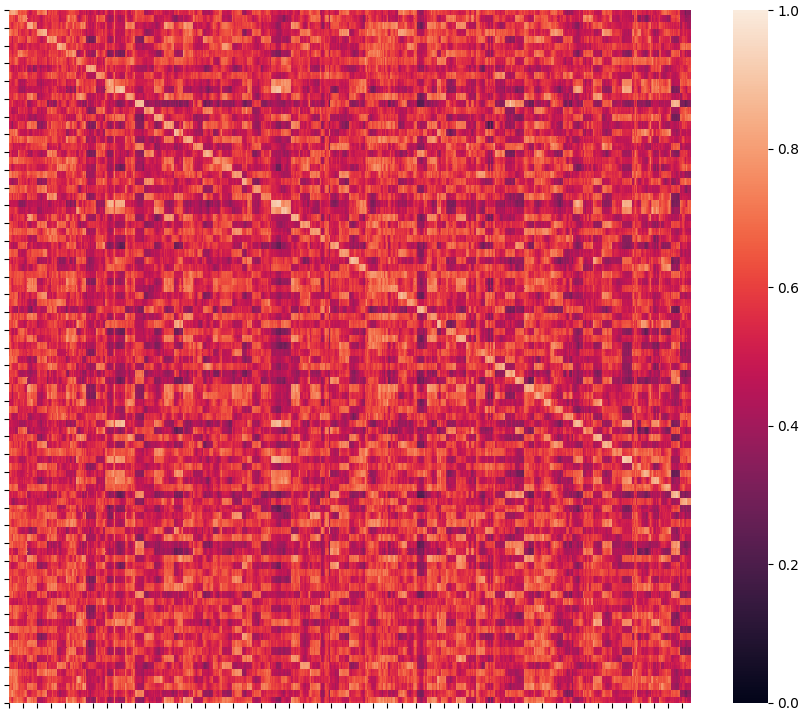}}
\caption{Cosine similarity score matrices using P\_TP0/G\_VB0 protocol of ARL-VTF. Horizontal and vertical axes represent probe and gallery imagery respectively.}
\label{fig:score}
\end{figure*}

\begin{figure}[htb]
\includegraphics[width=0.45\textwidth]{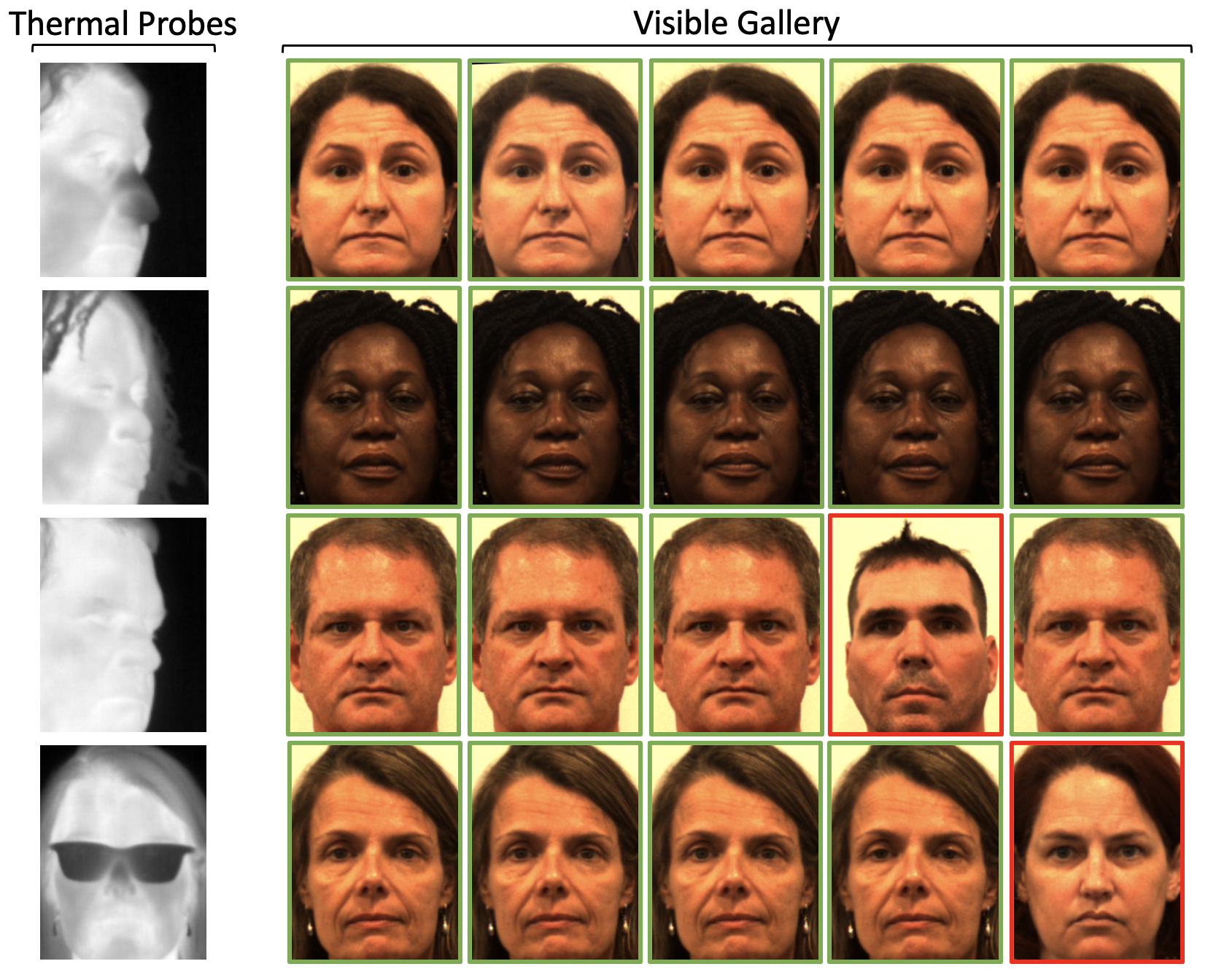}
\caption{Top-5 gallery match for thermal probes with pose and glasses conditions using proposed DPIF on thermal probes from ARL-VTF dataset.}
\label{fig:top5_results}
\end{figure}
 
\subsection{ARL Multimodal Face Dataset}
The ARL-MMF dataset Volume III is composed of over 5000 paired thermal and visible images collected at a single range ($2.5m$) from 126 subjects. These images include baseline, varying expressions, as well as varying pose from $-60^\circ$ to $+60^\circ$. Similar to~\cite{9304937}, we chose thermal off-pose and visible frontal (including baseline and expression) face pairs in our experiments and we also considered face imagery from 96 random subjects as the training set and face imagery from the remaining 30 subjects as the test set.

\subsection{Tufts Face Database}
This dataset is a multi-modal dataset with over 10,000 face images with pose and expressions variations from 112 subjects from multiple countries. For our experiments, we only considered a collection of 1532 paired thermal and visible face images. For any given subject, there are 9 images with pose variations (from $-90^\circ$ to $+90^\circ$), 4  with varying expressions and 1 with eyewear. This dataset is challenging as it does not provide as many off-pose imagery compare to the ARL-VTF and ARL-MMF datasets. For training our framework we chose images from 90 random subjects and the remaining 22 subjects were used for evaluation. So, we have 1232 paired images in the training set and 300 paired images in the testing set with no subject overlap across the two sets. 

\section{Experimental Results}
\label{sec:experiment}

In this section, we present the experiments we conducted on the datasets presented above as well as their respective results to show the effectiveness of our proposed framework. Particularly, we use the ARL-VTF dataset to evaluate our method on baseline, expression, and pose conditions and we use the ARL-MMF and Tufts dataset to evaluate our method on pose conditions to show effectiveness of the pose-correction across different datasets. For each dataset, we compare our proposed method's performance to some recent state-of-the-art approaches. In particular, the metrics we used to evaluate cross-spectrum performance across the different methods are typical face verification metrics associated with receiver operating characteristic (ROC) curves, including Area Under the Curve (AUC), Equal Error Rate (EER), and two additional points: True Accept Rate at 1\% and 5\% False Acceptance Rate--denoted TAR@1\%FAR and TAR@5\%FAR, respectively.  While these metrics can be computed using a variety of distance or similarity measures, we use cosine similarity to compute the reported metrics for our DPIF. 

Additionally, we perform a few ablation studies to understand (1) the impact of embedding size has on thermal-to-visible FR and the effect model capacity has on performance, (2) the effects of the hyperparameter $\lambda$ and the relative importance of the proposed pose-correction loss for cross-spectrum matching, and (3) the effect of truncating base architectures at various depth.

\subsection{Results on the ARL-VTF Dataset}

\begin{figure*}[htb]
\centering     %%% not \center
\subfloat[]{\label{fig:roc_pose_a}\includegraphics[width=0.50\textwidth]{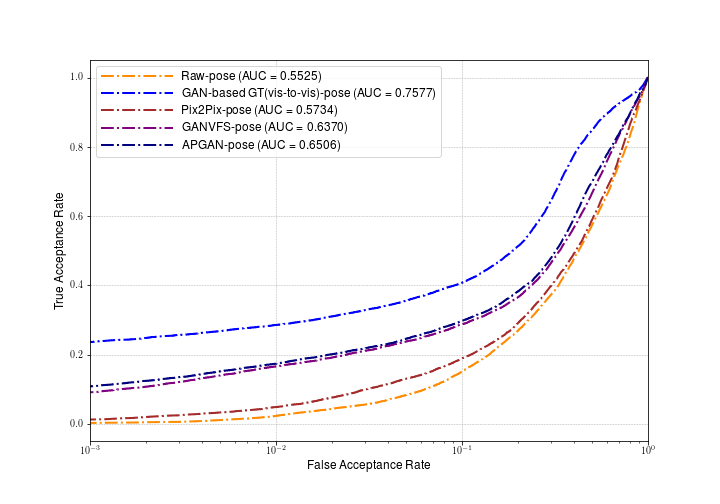}}
\subfloat[]{\label{fig:roc_pose_b}\includegraphics[width=0.50\textwidth]{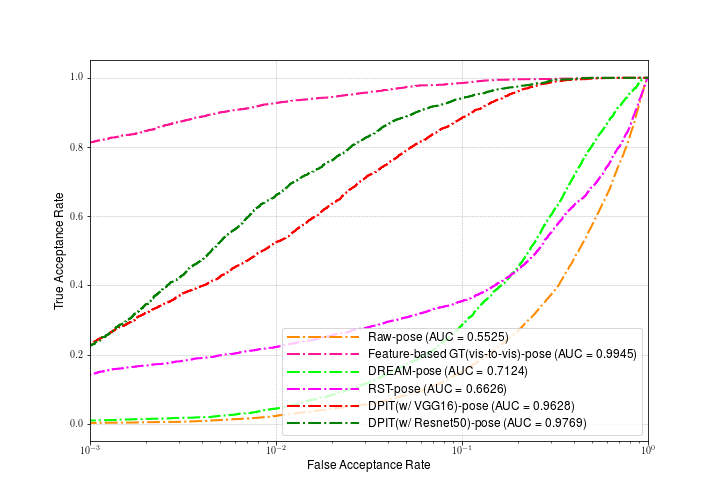}}
\caption{ROC curves for (a) image synthesis and (b) invariant image representation methods applied on P\_TP0/G\_VB0- using $pose$ condition}
\label{fig:roc_pose}
\end{figure*}

\begin{figure}[htb]
\includegraphics[width=0.50\textwidth]{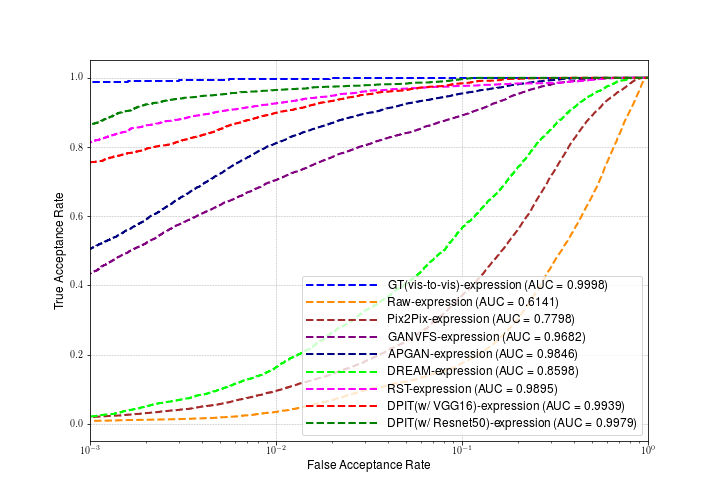}
\caption{ROC curves of the different methods applied on P\_TE0/G\_VB0- using $expression$ condition}
\label{fig:roc_expression}
\end{figure}

Using this dataset, we demonstrate efficacy of the proposed DPIF by showing both qualitative and quantitative results. In Fig.~\ref{fig:score}, we show visual comparison between ground-truth scoring matrix and other scoring matrices resulted from matching off-pose thermal probes to frontal visible gallery using  raw image representations, Dream~\cite{cao2018Dream}, RST~\cite{9304937}, and proposed DPIT. From this figure, we see that our proposed domain adaptation framework provides better matching scores compared to the other methods. 
We also show in Fig.~\ref{fig:top5_results} the top-5 images retrieved from the gallery after feeding thermal probes imagery with pose and glasses condition to the proposed framework. In this figure, the top two probe subjects are correctly matched against the gallery while the last two subjects present some false positives (i.e., subject mismatch) due to extremely challenging conditions exhibited by the probe imagery (e.g., off-pose and glasses) which differed from the frontal visible gallery imagery.

Additionally, we compare both our VGG16-based and Resnet50-based proposed frameworks with other recent competing methods such as pix2pix~\cite{pix2pix2017}, GANVFS~\cite{8272687}, DREAM~\cite{cao2018Dream},  DAL-GAN~\cite{XingHFF}, RST~\cite{9304937}, Axial-GAN~\cite{9484353}, and AP-GAN~\cite{Di2018PolarimetricTT}. While the approach used for the RST and DREAM models are similar to ours in the sense that they are both feature-based domain adaptation models, the others are based on Generative Adversarial Networks (GANs). As a baseline, we also include the results obtained from thermal-to-visible matching when using raw features extracted from the thermal probes and the visible gallery. Note that these raw image representations were extracted across different conditions and protocols using a similar feature extraction network as in~\cite{Poster_2021_WACV}. 

\begin{figure}[tb]
%\centering
\includegraphics[width=0.50\textwidth]{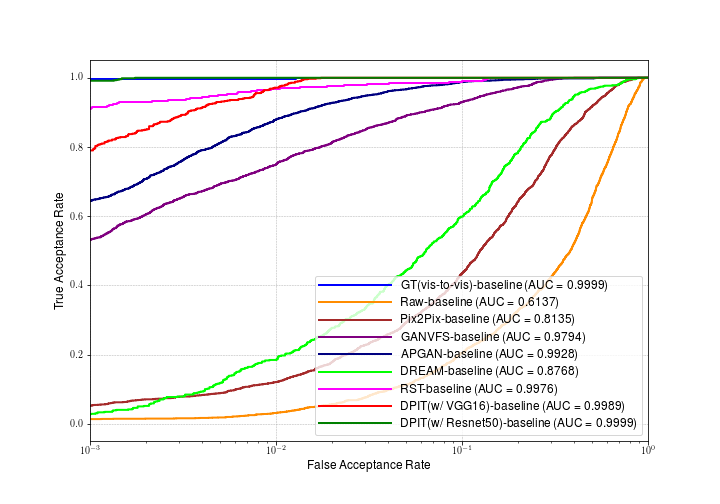}
\caption{ROC curves of the different methods applied on P\_TB0/G\_VB0- using $baseline$ condition}
\label{fig:roc_baseline}
\end{figure}

Tables \ref{tab:pose_table}, \ref{tab:expr_table}, and \ref{tab:base_table} illustrate the thermal-to-visible face verification performance across different methods on the ARL-VTF dataset respectively for pose, expression, and baseline conditions. Compared with the other state-of-the-art models shown in these tables, both our proposed Resnet50-based model and VGG16-based models perform better with higher AUC, TAR, and lower EER for baseline and pose conditions. While the RST method performs better than the proposed VGG16-based framework on the probes P\_TE0 in the expression case as shown in Table \ref{tab:expr_table}, the proposed Resnet50-based framework still provides the best performance across the different probe sets for the expression condition.
Additionally, we show on Fig.~\ref{fig:roc_pose}, Fig.~\ref{fig:roc_expression}, and Fig.~\ref{fig:roc_baseline} the ROC curves across various methods respectively for pose, expression and baseline conditions using gallery and probe sets from the ARL-VTF dataset. As demonstrated on these figures, our proposed method provides the best AUC compared to other domain adaptation models. Especially with the pose conditions, our proposed Domain and Pose Invariant Framework presents the highest AUC compared to other frontalization and generative models as shown on Fig.~\ref{fig:roc_pose_a} and Fig.~\ref{fig:roc_pose_b}.

Most recently, Peri et al.~\cite{9666943} made improvements on the ARL-VTF dataset using Contrastive Unpaired Translation (CUT)~\cite{park2020cut} along with several additional regularizing loss functions: alignment, L1, identity, and arc face~\cite{8953658}. While performance of individual conditions were not reported, the proposed CUT method in \cite{9666943} achieved 97.7\% AUC, 6.9\% EER, 77.2\% TAR@1\%FAR, and 90.5\% TAR@5\%FAR on average across baseline, expression, and pose conditions.  However, on average across all conditions, we achieved 98.99\% AUC, 3.38\% EER, 88.46\% TAR@1\%FAR, and 95.41\% TAR@5\%FAR.  Therefore, we still achieve the best performance across all conditions.  

Note that performance is saturated at (or near) 100\% for expression and baseline conditions (Table \ref{tab:expr_table} and Table \ref{tab:base_table}, respectively), which implies that future collections may need to acquire frontal imagery under more challenging conditions (e.g., at a distance).

\subsection{Results on the ARL-MMF Dataset}
Table \ref{tab:protocol3} shows the face verification performance of the proposed DPIF and other recent methods matching off-pose thermal probes with frontal visible gallery using volume III of the ARL-MMF dataset. Note that for this dataset, we did not retrain the proposed framework from scratch. Instead, we fine-tuned the proposed Resnet50-based model using the pretrained weights from the ARL-VTF dataset training and we report the mean performance along with the standard deviation across all training protocols. The results presented in this table demonstrate the advantage of the proposed method over the others as it offers the best performance. Note that this dataset is a more challenging dataset compare to the ARL-VTF dataset as it is about 100x smaller providing fewer images per subjects. This explains the drop in performance observed when evaluating the proposed domain adaptation framework on the ARL-MMF dataset Volume III compare to ARL-VTF performance. However, compared to previous state-of-the-art methods, our proposed DPIF still achieves improvements of 17.7\% and 24.99\% for TAR@1\%FAR and TAR@5\%FAR, respectively.

\begin{table*}[htb]
  \begin{center}
    \caption{Face verification results using $pose$ condition from ARL-VTF Dataset}
    \label{tab:pose_table}
    \begin{tabular}{lc*{11}{c}}
        \toprule 
        \multicolumn{1}{c}{\multirow{2}{*}{Probes}} & \multicolumn{1}{c}{\multirow{2}{*}{Method}} & \multicolumn{4}{c}{\textbf{Gallery G\_VB0-}} & \multicolumn{1}{c}{\multirow{2}{*}{}} & \multicolumn{4}{c}{\textbf{Gallery G\_VB0+}}\\ \cmidrule{3-6} \cmidrule{8-11} 
        \multicolumn{2}{c}{}  & AUC(\%) & EER(\%) & TAR@1\%FAR & TAR@5\%FAR &  & AUC(\%) & EER(\%) & TAR@1\%FAR & TAR@5\%FAR \\ \midrule
        
        & Raw & 55.24 & 46.25 & 2.23 & 8.25 & & 55.10 & 46.34 & 2.91 & 8.74 \\ 
        & Pix2Pix~\cite{pix2pix2017} & 54.86 & 47.22 & 3.13 & 9.78 & & 56.50 & 46.03 & 4.01 & 10.84 \\ 
        & GANVFS~\cite{8272687} & 63.70 & 41.66 & 16.55 & 23.73 & & 65.58 & 40.19 & 17.95 & 25.68 \\ 
        \textbf{P\_TP0} & AP-GAN~\cite{Di2018PolarimetricTT} & 65.06 & 40.24 & 17.33 & 24.56 & & 67.13 & 38.67 & 18.91 & 26.46\\  
        & RST~\cite{9304937} & 66.26 & 38.05 & 22.18 & 30.72 & & 68.39 & 36.86 & 22.64 & 31.81\\
        & DREAM~\cite{cao2018Dream} & 71.24 & 34.60 & 4.36 & 17.07 & & 75.87 & 31.20 & 6.46 & 21.43\\
        & DAL-GAN~\cite{XingHFF} & 77.48 & 29.08 & 8.20 & 25.89 & & - & - & - & - \\ 
        & VGG16 + DPIT (Ours) & 96.28 & 10.80 & 52.37 & 78.86 & & 96.27 & 10.14 & 46.76 & 78.01\\
        & Resnet50 + DPIT (Ours) & \textbf{97.69} & \textbf{7.75} & \textbf{66.08} & \textbf{88.74} & & \textbf{97.39} & \textbf{8.39} & \textbf{68.00} & \textbf{88.54} \\
        \bottomrule
        
        & Raw & 55.48 & 45.98 & 3.25 & 8.47 & & 56.82 & 44.74 & 2.09 & 7.57 \\ 
        & Pix2Pix~\cite{pix2pix2017} & 54.31 & 47.04 & 2.93 & 8.44 & & 50.08 & 49.67 & 0.60 & 4.33 \\ 
        & GANVFS~\cite{8272687} & 65.79 & 40.35 & 17.84 & 25.48 & & 59.51 & 44.04 & 4.29 & 15.47 \\ 
        \textbf{P\_TP-} & AP-GAN~\cite{Di2018PolarimetricTT} & 67.27 & 39.00 & 18.16 & 26.02 & & 60.10 & 43.57 & 5.77 & 15.97 \\ 
        & RST~\cite{9304937} & 68.24 & 37.60 & 23.09 & 33.54 & & 63.29 & 41.79 & 18.79 & 27.93\\
        & DREAM~\cite{cao2018Dream} & 73.63 & 33.33 & 9.25 & 23.47 & & 62.93 & 41.47 & 2.53 & 8.93\\
        & DAL-GAN~\cite{XingHFF} & 82.18 & 25.11 & 10.82 & 30.64 & & - & - & - & -\\ 
        & VGG16 + DPIT (Ours) & 96.46 & 10.33 & 56.68 & 79.93 & & 94.44 & 12.54 & 35.11 & 67.75\\
        & Resnet50 + DPIT (Ours) & \textbf{97.09} & \textbf{9.09} & \textbf{63.91} & \textbf{84.40} & & \textbf{96.62} & \textbf{10.39} & \textbf{55.84} & \textbf{78.73}\\
        \bottomrule
    \end{tabular}
  \end{center}
\end{table*}

\begin{table*}[htb]
  \begin{center}
    \caption{Face verification results using $expression$ condition from ARL-VTF Dataset}
    \label{tab:expr_table}
    \begin{tabular}{lc*{11}{c}}
        \toprule 
        \multicolumn{1}{c}{\multirow{2}{*}{Probes}} & \multicolumn{1}{c}{\multirow{2}{*}{Method}} & \multicolumn{4}{c}{\textbf{Gallery G\_VB0-}} & \multicolumn{1}{c}{\multirow{2}{*}{}} & \multicolumn{4}{c}{\textbf{Gallery G\_VB0+}}\\ \cmidrule{3-6} \cmidrule{8-11} 
        \multicolumn{2}{c}{}  & AUC(\%) & EER(\%) & TAR@1\%FAR & TAR@5\%FAR &  & AUC(\%) & EER(\%) & TAR@1\%FAR & TAR@5\%FAR \\ \midrule
        
        & Raw & 61.40 & 41.96 & 3.40 & 12.18 & & 62.50 & 41.38 & 4.60 & 13.25 \\ 
        & Pix2Pix~\cite{pix2pix2017} & 69.10 & 35.98 & 7.01 & 16.44 & & 73.97 & 31.87 & 7.93 & 19.60 \\ 
        & DREAM~\cite{cao2018Dream} & 85.98 & 22.54 & 16.36 & 40.46 & & 86.62 & 21.72 & 18.47 & 42.49\\
        & GANVFS~\cite{8272687} & 96.81 & 10.51 & 70.41 & 84.00 & & 97.73 & 8.90 & 74.20 & 86.80 \\ 
        \textbf{P\_TE0} & AP-GAN~\cite{Di2018PolarimetricTT} & 98.46 & 6.44 & 81.11 & 92.49 & & 98.89 & 5.60 & 84.23 & 93.94\\ 
        & RST~\cite{9304937} & 98.95 & 3.61 & 92.61 & 96.88 & & 99.01 & 3.57 & 92.69 & 96.93\\
        & VGG16 + DPIT (Ours) & 99.39 & 4.01 & 89.89 & 96.46 & & 99.58 & 3.19 & 93.03 & 98.09\\
        & Resnet50 + DPIT (Ours) & \textbf{99.79} & \textbf{2.39} & \textbf{96.49} & \textbf{98.31} & & \textbf{99.70} & \textbf{2.33} & \textbf{96.52} & \textbf{98.29}\\
        \bottomrule
        
        & Raw & 63.26 & 42.34 & 4.66 & 16.28 & & 59.33 & 43.17 & 2.04 & 8.00 \\ 
        & Pix2Pix~\cite{pix2pix2017} & 68.78 & 36.24 & 7.75 & 18.06 & & 51.05 & 49.11 & 2.26 & 4.95 \\ 
        & DREAM~\cite{cao2018Dream} & 87.01 & 22.06 & 21.06 & 41.07 & & 72.42 & 35.07 & 8.60 & 21.13\\
        & Axial-GAN~\cite{9484353} & 92.71 & 14.86 & - & - & & 88.01 & 21.58 & - & - \\ 
        & GANVFS~\cite{8272687} & 98.66 & 5.93 & 73.17 & 92.82 & & 83.68 & 22.41 & 6.77 & 22.13 \\ 
        \textbf{P\_TE-} & AP-GAN~\cite{Di2018PolarimetricTT} & 99.30 & 3.84 & 82.55 & 97.44 & & 86.12 & 21.68 & 9.88 & 31.62\\ 
        & RST~\cite{9304937} & 99.83 & 2.27 & 95.66 & 99.48 & & 99.48 & 3.05 & 89.45 & 98.07\\
        & VGG16 + DPIT (Ours) & \textbf{99.90} & 1.72 & 97.19 & 99.80 & & 99.37 & 3.81 & 89.66 & 96.39\\
        & Resnet50 + DPIT (Ours) & 99.88 & \textbf{0.81} & \textbf{99.47} & \textbf{99.87} & & \textbf{99.77} & \textbf{2.92} & \textbf{95.33} & \textbf{98.87}\\
        \bottomrule
    \end{tabular}
  \end{center}
\end{table*}

\begin{table*}[htb]
  \begin{center}
    \caption{Face verification results using $baseline$ condition from ARL-VTF Dataset}
    \label{tab:base_table}
    \begin{tabular}{lc*{11}{c}}
        \toprule 
        \multicolumn{1}{c}{\multirow{2}{*}{Probes}} & \multicolumn{1}{c}{\multirow{2}{*}{Method}} & \multicolumn{4}{c}{\textbf{Gallery G\_VB0-}} & \multicolumn{1}{c}{\multirow{2}{*}{}} & \multicolumn{4}{c}{\textbf{Gallery G\_VB0+}}\\ \cmidrule{3-6} \cmidrule{8-11} 
        \multicolumn{2}{c}{}  & AUC(\%) & EER(\%) & TAR@1\%FAR & TAR@5\%FAR &  & AUC(\%) & EER(\%) & TAR@1\%FAR & TAR@5\%FAR \\ \midrule
        
        & Raw & 61.37 & 43.36 & 3.13 & 11.28 & & 62.83 & 42.37 & 4.19 & 13.29 \\ 
        & Pix2Pix~\cite{pix2pix2017} & 71.12 & 33.80 & 6.95 & 21.28 & & 75.22 & 30.42 & 8.28 & 27.63 \\
        & DREAM~\cite{cao2018Dream} & 87.68 & 20.58 & 18.56 & 44.36 & & 89.21 & 18.82 & 21.15 & 45.61\\
        & GANVFS~\cite{8272687} & 97.94 & 8.14 & 75.00 & 88.93 & & 98.58 & 6.94 & 79.09 & 91.04 \\ 
        \textbf{P\_TB0} & AP-GAN~\cite{Di2018PolarimetricTT} & 99.28 & 3.97 & 87.95 & 96.66 & & 99.49 & 3.38 & 90.52 & 97.81\\ 
        & RST~\cite{9304937} & 99.76 & 2.30 & 96.84 & 98.43 & & 99.87 & 1.84 & 97.29 & 98.80\\
        & VGG16 + DPIT (Ours) & 99.89 & 1.20 & 97.19 & \textbf{100.00} & & 99.84 & 2.12 & 94.89 & \textbf{100.00}\\
        & Resnet50 + DPIT (Ours) & \textbf{99.99} & \textbf{0.15} & \textbf{100.00} & \textbf{100.00} & & \textbf{100.00} & \textbf{0.12} & \textbf{100.00} & \textbf{100.00}\\
        \bottomrule
        
        & Raw & 61.14 & 41.64 & 2.77 & 16.11 & & 57.61 & 44.73 & 1.38 & 6.11 \\ 
        & Pix2Pix~\cite{pix2pix2017} & 68.77 & 38.02 & 6.69 & 20.28 & & 52.11 & 48.88 & 2.22 & 4.66 \\ 
        & DREAM~\cite{cao2018Dream} & 87.48 & 20.01 & 14.78 & 39.71 & & 74.39 & 32.02 & 4.23 & 20.92\\
        & Axial-GAN~\cite{9484353} & 94.4 & 12.38 & - & - & & 89.71 & 19.75 & - & - \\ 
        & GANVFS~\cite{8272687} & 99.36 & 3.77 & 84.88 & 97.66 & & 87.34 & 18.66 & 7.00 & 29.66 \\ 
        \textbf{P\_TB-} & AP-GAN~\cite{Di2018PolarimetricTT} & 99.63 & 2.66 & 91.55 & 98.88 & & 89.24 & 19.49 & 16.33 & 41.22\\ 
        & RST~\cite{9304937} & 99.83 & 1.95 & 96.00 & 99.48 & & 99.03 & 4.79 & 85.56 & 95.86\\
        & VGG16 + DPIT (Ours) & 99.94 & 1.01 & 99.29 & \textbf{100.00} & & \textbf{99.82} & 1.60 & 95.10 & \textbf{100.00}\\
        & Resnet50 + DPIT (Ours) & \textbf{100.00} & \textbf{0.00} & \textbf{100.00} & \textbf{100.00} & & 97.97 & \textbf{0.66} & \textbf{100.00} & \textbf{100.00}\\
        \bottomrule
        
        & Raw & 59.52 & 42.60 & 4.66 & 6.00 & & 78.26 & 29.77 & 3.88 & 21.33 \\
        & Pix2Pix~\cite{pix2pix2017} & 59.68 & 41.72 & 3.33 & 3.33 & & 67.08 & 36.44 & 2.68 & 11.11 \\
        & DREAM~\cite{cao2018Dream} & 74.63 & 29.97 & 3.84 & 23.05 & & 75.49 & 32.28 & 8.85 & 29.72\\
        & Axial-GAN~\cite{9484353} & 84.62 & 24.67 & - & - & & 93.62 & 14.05 & - & - \\ 
        & GANVFS~\cite{8272687} & 87.61 & 20.16 & 20.55 & 44.66 & & 96.82 & 8.66 & 46.77 & 83.00 \\ 
        \textbf{P\_TB+} & AP-GAN~\cite{Di2018PolarimetricTT} & 91.11 & 17.43 & 22.33 & 55.66 & & 97.96 & 7.21 & 60.11 & 88.70\\ 
        & RST~\cite{9304937} & 99.28 & 5.32 & 89.21 & 94.79 & & 99.97 & 0.73 & 99.47 & \textbf{100.00}\\
        & VGG16 + DPIT (Ours) & 99.61 & 3.40 & 92.46 & 98.81 & & 99.95 & 0.70 & \textbf{100.00} & \textbf{100.00}\\
        & Resnet50 + DPIT (Ours) & \textbf{99.91} & \textbf{1.94} & \textbf{96.84} & \textbf{100.00} & & \textbf{100.00} & \textbf{0.32} & \textbf{100.00} & \textbf{100.00}\\
        \bottomrule
    \end{tabular}
  \end{center}
\end{table*}

\begin{table*}[htb]
\begin{center}
\caption{Face verification results on the ARL-MMF Dataset}
\label{tab:protocol3}
\begin{tabular}{ |P{1.7cm}|P{2.5cm}|P{1.7cm}|P{1.7cm}|P{1.7cm}|P{1.7cm}|}
\hline
\textbf{Conditions} & \textbf{Methods} & \textbf{AUC (\%)} & \textbf{EER (\%)} & \textbf{TAR@1\%FAR (\%)} & \textbf{TAR@5\%FAR (\%)} \\
\hline
\multirow{9}{*}{Pose} & Raw & 64.12 & 40.51 & 4.21 & 17.74 \\
& RST~\cite{9304937} & 63.02 & 42.23 & 9.93 & 22.14\\ 
& Pix2pix~\cite{pix2pix2017} & 73.60 & 31.96 & 11.16 & 26.45\\
& TP-GAN~\cite{tpgan} & 76.15 & 30.89 & 6.26 & 19.03\\
& PIM~\cite{8578333} & 80.89 & 26.86 & 9.83 & 27.11\\
& Multi-AP-GAN~\cite{9358101} & 82.35 & 25.76 & - & -\\
& DA-GAN~\cite{9320279} & 86.26 & 21.56 & 14.74 & 33.17\\
& DAL-GAN~\cite{XingHFF} & 88.61 & 20.21 & 16.07 & 40.93\\
& DPIT(Ours) & \textbf{93.06 $\pm$ 0.42} & \textbf{14.97 $\pm$ 0.44} & \textbf{33.77 $\pm$ 4.94} & \textbf{64.92 $\pm$ 2.08} \\  
\hline
\end{tabular}
\end{center}
\end{table*}

\begin{table*}[htb]
\begin{center}
\caption{Face verification results on the Tufts Face Dataset }
\label{tab:Tufts}
\begin{tabular}{ |P{1.7cm}|P{1.8cm}|P{1.7cm}|P{1.7cm}|P{1.7cm}|P{1.7cm}|}
\hline
\textbf{Conditions} & \textbf{Methods} & \textbf{AUC (\%)} & \textbf{EER (\%)} & \textbf{TAR@1\%FAR (\%)} & \textbf{TAR@5\%FAR (\%)} \\
\hline
\multirow{7}{*}{Pose} & Raw & 67.55 & 37.88 & 5.11 & 16.11 \\
& RST~\cite{9304937} & 65.63 & 40.01 & 10.51 & 21.69\\ 
& Pix2pix~\cite{pix2pix2017} & 69.71 & 35.31 & 5.44 & 21.66\\
& TP-GAN~\cite{tpgan} & 70.93 & 35.32 & 6.46 & 18.77\\
& PIM~\cite{8578333} & 72.84 & 34.10 & 8.77 & 21.00\\
& DA-GAN~\cite{9320279} & 75.24 & 31.14 & 10.44 & 26.22\\
& DAL-GAN~\cite{XingHFF} & 78.68 & 28.38 & 10.44 & 27.11\\
& CUT-ATC~\cite{9666943} & 87.4 & 21.3 & 26.2 & 50.6\\
& DPIT(Ours) & \textbf{91.82 $\pm$ 0.35} & \textbf{16.03$\pm$ 0.53} & \textbf{31.59 $\pm$ 2.64} & \textbf{58.75 $\pm$ 1.83} \\  
\hline
\end{tabular}
\end{center}
\end{table*}

\subsection{Results on the Tufts Face Dataset}
Table \ref{tab:Tufts} shows the cross-spectrum face verification performance of several methods including the proposed DPIF on the Tufts dataset using for the pose condition. Similar to the ARL-MMF dataset, we fine-tuned the proposed Resnet50-based network using the same pretrained weights from the ARL-VTF dataset training and report the mean performance along with the standard deviation across all training protocols. With this other challenging dataset, the proposed method surpasses the competing ones by improving state-of-the-art performance by 11.15\%, 10.88\% and 12.44\% in AUC, EER, and TAR@1\%FAR, respectively. Note that we get relatively lower performance with the Tufts dataset as this dataset was collected with older thermal cameras that had lower resolution (i.e., $336 \times 256$ for Tufts compared to $640 \times 512$ for ARL-VTF) and lower sensitivity. This, along with the relative small size of the Tufts dataset do not provide as much learnable information during training as the ARL-VTF dataset.

\subsection{Ablation Studies}
\label{sec:ablation}
\textbf{Embedding Size:} Table~\ref{tab:embedding} shows the impact of the feature vector length on profile-to-frontal cross-spectrum FR. As shown in the table, the proposed DPIF is more discriminative with a larger feature vector until performance saturates between 256- and 512- dimensional representations. However, peak performance is obtained at the cost of a relatively large number of trainable parameters which increases computation complexity, memory requirements, and vulnerability to over-fitting. In fact, it could be seen that as the image representations vector doubles in size the number of trainable parameters increases by a factor of about 1.8. Our proposed method uses 256 as the optimal embedding size since larger feature vectors do not provide a significant improvement and the number of trainable parameters almost doubles and quadruples respectively for an embedding size of 512 and 1024, respectively.
\begin{table*}[htb]
\begin{center}
\caption{Ablation study on the effect of embedding size on performance.}
\label{tab:embedding}
\begin{tabular}{ |P{1.5cm}|P{1.5cm}|P{1cm}|P{1cm}|P{1.7cm}|P{1.7cm}|}
\hline
\textbf{Embedding Size} & \textbf{Trainable Parameters} & \textbf{AUC (\%)} & \textbf{EER (\%)} & \textbf{TAR@1\%FAR (\%)} & \textbf{TAR@5\%FAR (\%)} \\
\hline
64 & 4.5M & 94.92 & 12.69 & 45.63 & 73.89 \\
128 & 7.7M & 96.58 & 10.20 & 54.27 & 82.60\\
256 & 14.1M & 97.69 & 7.75 & 66.08 & 88.74\\
512 & 26.9M & 97.68 & 7.83 & 67.54 & 88.29\\
1024 & 52.6M & 97.51 & 8.18 & 66.82 & 87.72\\
\hline
\end{tabular}
\end{center}
\end{table*}

\begin{figure}
\captionsetup{justification=centering}
\includegraphics[width=0.45\textwidth]{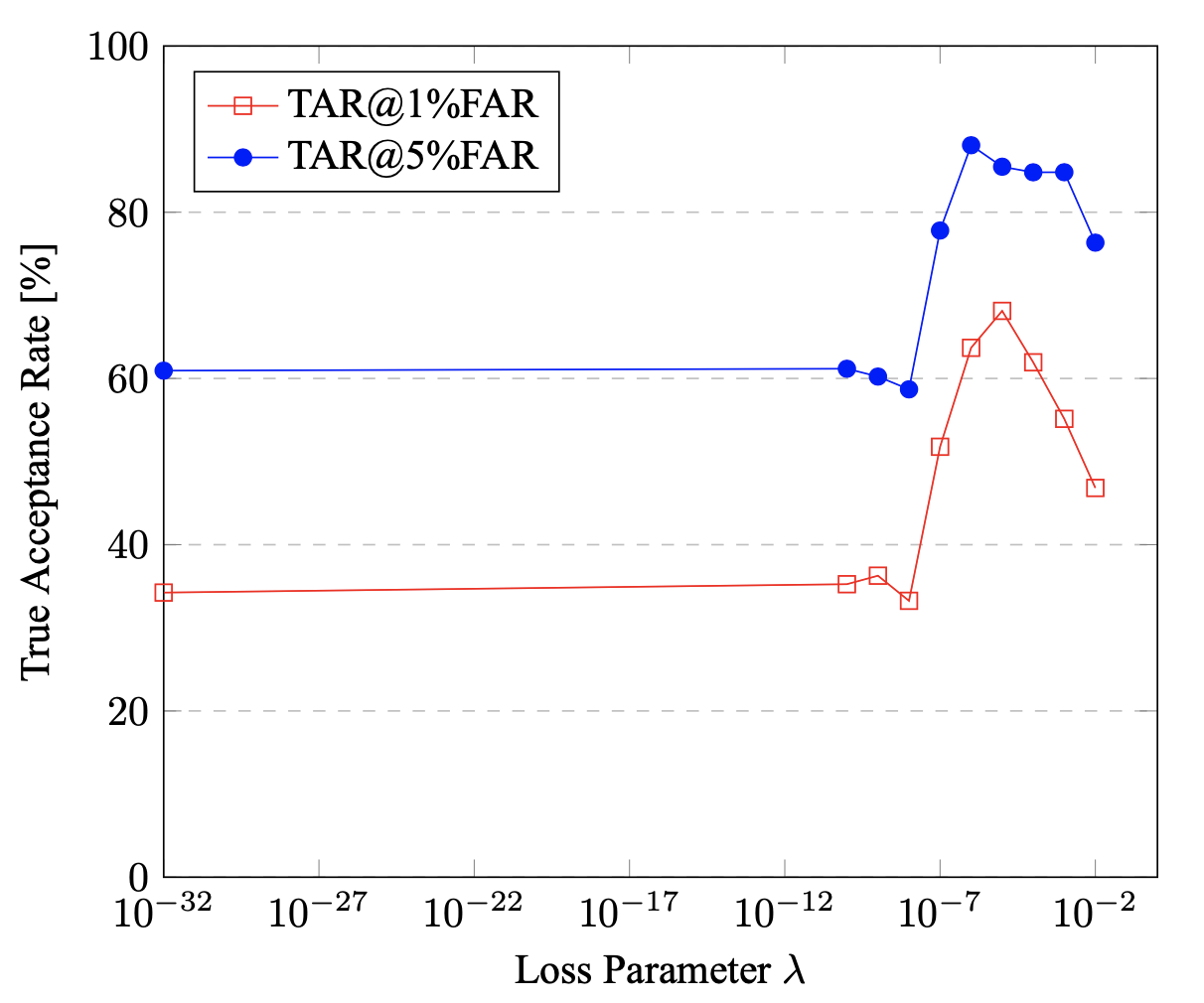}
\caption{Ablation study on the effect of pose correction loss on performance by varying $\lambda$.}
\label{fig:loss_param}
\end{figure}

\textbf{Loss Parameter:} We show in Fig.~\ref{fig:loss_param} how the loss parameter $\lambda$ effects True Acceptance Rate (TAR) performances. We observe highest TAR performance when $\lambda=10^{-5}$ and when $\lambda=10^{-6}$ at 1\%FAR and 5\%FAR, respectively. As mentioned in the methodology section, we use $\lambda=10^{-5}$ as the optimal loss parameter since it yields highest TAR@1\%FAR and thus provides minimal false positives during face verification. Note that when $\lambda=0$ (i.e., no pose-correction loss), the joint loss function described in Eq.~\ref{eq:total_loss} simply becomes the cross-spectrum loss. In a such situation, we get 34.24\% and 60.94\% for TAR@1\%FAR and TAR@5\%FAR, respectively, which is significantly low compared to the peak performance as shown in Fig.~\ref{fig:loss_param}. This suggests that training the DPIF using only the proposed cross-spectrum loss is not sufficient to bridge pose and domain gap. This demonstrates effectiveness of the proposed joint-loss function in learning robustness to pose and domain variations. Although the DPIF trained solely with proposed cross-spectrum loss provides lower performance, it still achieves around 12.06\% and 30.22\% improvements in TAR at 1\%FAR and 5\%FAR respectively over the approach described in~\cite{9304937}.

\textbf{Modified Base Architecture Output Feature Maps:} As illustrated in Table~\ref{tab:feat_extract}, both base architectures are composed of five main convolutional blocks. We herein study the effect of truncating each base architecture at various depths on off-pose to frontal cross-spectrum matching performance. In Table~\ref{tab:feat_map}, we present for each base architecture, feature map dimensions at different blocks with the corresponding Equal Error Rate (EER), and True Accept Rate at 1\% when matching thermal off-pose face imagery with visible frontal face imagery from ARL-VTF dataset. Note that these results were obtained by matching off-pose thermal and frontal visible image embeddings using cosine similarity. While~\cite{9304937} proposed to truncate Resnet50 and VGG16 networks respectively at `block3\_pool' and `block\_3d' for better thermal-to-visible frontal face matching, the results shown in this table rather suggest that better thermal off-pose to visible frontal matching is achieved when truncating base architectures right after `block4\_pool' and `block4\_e' for VGG16 and Resnet50, respectively. This indicates that although early intermediate layers (i.e., `block3\_pool' and `block\_3d') provide enough similarity between frontal thermal and frontal visible image representations, the pose gap remains significant when matching thermal off-pose to visible frontal face imagery. Interestingly, deeper intermediate layers (i.e., `block4\_pool' and `block4\_e') not only provide images representations that are less dependent on pose and spectrum, but they also offer sufficient contextual information for matching thermal off-pose probes to visible frontal gallery.

\begin{table}[htb]
\caption{Modified base architectures performance per output feature maps}
\label{tab:feat_map}
\resizebox{\columnwidth}{!}{%
\begin{tabular}{ |P{1.2cm}|P{1.5cm}|P{1.7cm}|P{1cm}|P{1.7cm}|}
\hline
\textbf{Base Network} & \textbf{Block Names} & \textbf{Feature Maps ($H \times W \times C$)} &\textbf{EER (\%)} & \textbf{TAR@1\%FAR (\%)}\\
\hline
\multirow{4}{*}{VGG16} & Block2\_pool & $56 \times 56 \times 128$ & 47.91 & 0.40 \\
& Block3\_pool & $28 \times 28 \times 256$ & 46.31 & 2.03\\
& Block4\_pool & $14 \times 14 \times 512$ & 47.09 & \textbf{8.15}\\
& Block5\_pool & $7 \times 7 \times 512$ & 45.63 & 2.18\\
\hline
\multirow{4}{*}{ResNet50} & Block\_2c & $28 \times 28 \times 256$ & 46.37 & 2.11\\
& Block\_3d & $14 \times 14 \times 512$ & 41.54 & 4.00\\
& Block\_4e & $14 \times 14 \times 1024$ & 44.60 & \textbf{15.71}\\
& Block\_4f & $7 \times 7 \times 1024$ & 45.17 & 13.40\\
& Block\_5c & $7 \times 7 \times 2048$ & 45.49 & 2.65\\
\hline
\end{tabular}
}
\end{table}

\textbf{Activation Functions:} Table~\ref{tab:activation_func} shows how different combinations of activation functions in the DPIT block can impact performance when matching profile thermal imagery to frontal visible imagery. Each combination presented in the table consists of either \textit{tanh}, \textit{ReLu}, or both. Additionally, the first and the second activation functions correspond to the first and the second convolutional block of the DPIT, respectively. As shown in Table~\ref{tab:activation_func}, the proposed framework achieves best performance when training the proposed framework using \textit{tanh} in the 3-layer convolutional block and \textit{ReLu} in the 2-layer convolutional block of DPIT. Note that these results were obtained using the proposed truncated Resnet50 with an embedding size of 256.

\begin{table}[htb]
\caption{Ablation study on the effect of activation functions on performance.}
\label{tab:activation_func}
\resizebox{\columnwidth}{!}{%
\begin{tabular}{|P{1.5cm}|P{1cm}|P{1cm}|P{1.7cm}|P{1.7cm}|}
\hline
\textbf{Activation Functions} & \textbf{AUC (\%)} &\textbf{EER (\%)} & \textbf{TAR@1\%FAR (\%)} & \textbf{TAR@5\%FAR (\%)}\\
\hline
ReLu-ReLu & 96.37 & 10.43 & 56.99 & 83.09\\
ReLu-tanh & 96.42 & 9.46 & 58.22 & 82.11\\
tanh-tanh & 97.53 & 8.92 & 66.44 & 87.54\\
tanh-ReLu & \textbf{97.69} & \textbf{7.75} & \textbf{66.08} & \textbf{88.74}\\
\hline
\end{tabular}
}
\end{table}

\section{Conclusion}
\label{sec:conclusion}
In this work, we proposed and analyzed a novel Domain and Pose Invariance Framework (DPIF) for off-pose thermal to frontal visible face matching, which used modified base architectures to extract image representations along with a new sub-network to simultaneously learn pose and domain invariance using new joint-loss function that combines proposed pose-correction and cross-spectrum losses. Qualitative and quantitative results from experiments conducted on three thermal-to-visible face datasets: ARL-VTF, ARL-MMF, and Tufts Face demonstrate significant improvements over state-of-the-art domain adaptation methods not only on pose conditions, but also on baseline and varying expressions.  

Additionally, ablation studies were conducted to illustrate the effects of embedding size, pose-correction loss, and adapting pretrained networks for thermal-to-visible FR, especially for off-pose to frontal cross-spectrum matching. Most importantly, we have developed a relatively compact and efficient domain adaptation framework that (1) does not require a priori information relating to head pose of probe imagery and (2) differs from image-to-image translation approaches.  In fact, we demonstrated how our DPIF outperforms many image-to-image translation (or GAN) methods for thermal-to-visible FR.

\begin{appendices}
\section{Results on the CASIA NIR-VIS 2.0 Database}
\label{appendix:casia}
In this section we consider NIR-to-VIS face recognition using the CASIA NIR-VIS 2.0 dataset~\cite{6595898} which is a much smaller dataset in terms of number of images, but contains more subject variation than ARL-VTF. Table~\ref{tab:casia} shows comparison between the proposed DPIT and recent NIR-to-VIS methods such as SeetaFace~\cite{Liu2016VIPLFaceNetAO}, DLBP~\cite{7301308}, Gabor+RBM~\cite{7163093}, IDNet~\cite{7789537}, Hallucination~\cite{8100203}, CenterLoss~\cite{Wen2016ADF}, TRIVET~\cite{7550064}, DVG~\cite{Fu2019DualVG}. We compare these methods  by computing face identification (Rank-1) and face verification (TAR@0.1\%FAR) performance on CASIA NIR-VIS 2.0 dataset. While DPIT outperforms most methods presented in the table, TRIVET and DVG provide better identification and verification performance than DPIT. A key difference between these methods (i.e., DVG and TRIVET) and DPIT is that, DVG is a synthesis method which requires a relatively large amount of trainable parameters and TRIVET is pretrained on the CASIA WebFace dataset~\cite{14117923} (494,414 images corresponding 10,575 subjects) to address the problem of limited amount of paired NIR-VIS face images in CASIA NIR-VIS 2.0.

\begin{table}[ht]
\begin{center}
\caption{Face identification / verification results on CASIA NIR-VIS-2.0 dataset }
\label{tab:casia}
\begin{tabular}{|P{2.5cm}|P{2.5cm}|P{2.5cm}|}
\hline
\textbf{Methods} & \textbf{Rank-1 (\%)} & \textbf{TAR@0.1\%FAR (\%)} \\
\hline
Raw & 5.42 $\pm$ 0.84 & 1.65 $\pm$ 1.20 \\
SeetaFace~\cite{Liu2016VIPLFaceNetAO} & 68.03 $\pm$ 1.66 & 58.75 $\pm$ 2.26 \\
DLBP~\cite{7301308} & 78.46 $\pm$ 1.67 & 85.80 \\
Gabor+RBM~\cite{7163093} & 86.16 $\pm$ 0.98 & 81.29 $\pm$ 1.82\\
IDNet~\cite{7789537} & 87.10 $\pm$ 0.9 & 74.5 \\
CenterLoss~\cite{Wen2016ADF} & 87.69 $\pm$ 1.45 & 69.72 $\pm$ 2.07\\
Hallucination~\cite{8100203} & 89.60 $\pm$ 0.9 & - \\
\textbf{DPIT(Ours)} & \textbf{91.85 $\pm$ 0.73} & \textbf{89.60 $\pm$ 1.09} \\
TRIVET~\cite{7550064} & 95.7 $\pm$ 0.5 & 91.0 $\pm$ 1.3\\ 
DVG~\cite{Fu2019DualVG} & 99.2 $\pm$ 0.3 & 98.8 $\pm$ 0.3\\ 
\hline
\end{tabular}
\end{center}
\end{table}
\end{appendices}

% \section*{Acknowledgment}
% TODO

{\small
\bibliographystyle{ieee}
\bibliography{main}
}
\vspace{-0.45in}
\begin{IEEEbiography}[{\includegraphics[width=1in,height=1.25in,clip,keepaspectratio]{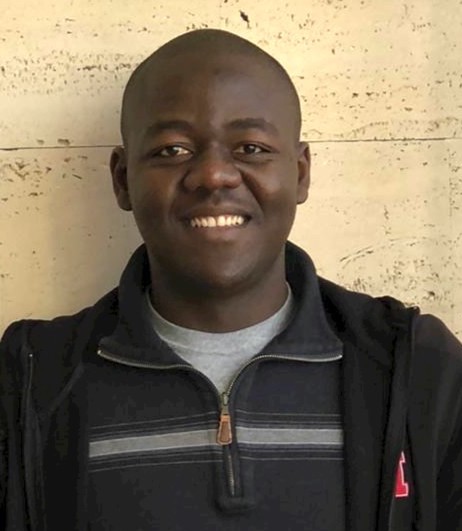}}]{Cedric Nimpa Fondje}{\space}(Student Member, IEEE) received the B.S. degree in electrical engineering from the University of Nebraska-Lincoln (UNL) in 2019. He was awarded the Nebraska Engineering Recruitment Fellowship and the Milton E. Mohr Fellowship in 2019 and 2021, respectively.  He is currently pursing his Ph.D. in electrical engineering from UNL with research interest in computer vision, biometrics, and domain adaptation.  He has published several papers on cross-spectrum face recognition. 
\end{IEEEbiography}
\vspace{-5.1in}
\begin{IEEEbiography}[{\includegraphics[width=1in,height=1.25in,clip,keepaspectratio]{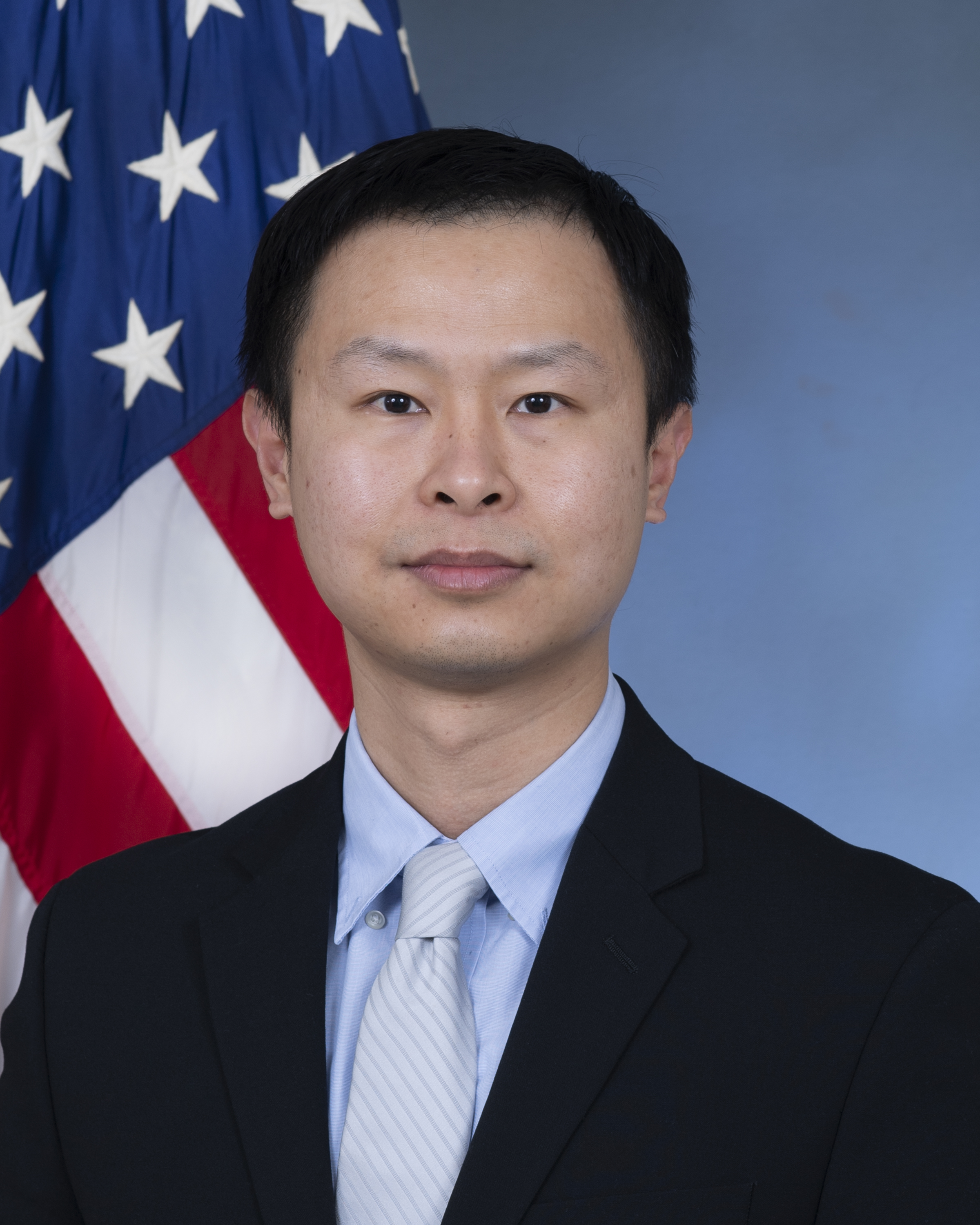}}]{Shuowen Hu}{\space}(Member, IEEE) received the B.S. degree in electrical and computer engineering from Cornell University, in 2005, and the Ph.D. degree in electrical and computer engineering from Purdue University, in 2009. He was awarded the Andrews Fellowship to study at Purdue University, conducting research in biomedical signal processing. Following graduation from Purdue University, he joined the DEVCOM Army Research Laboratory (ARL) as an Electronics Engineer with the Image Processing Branch. He has more than 50 conference and journal publications. His current research interests include cross-spectrum face recognition as well as on object detection and classification. 
\end{IEEEbiography}
\vspace{-5.0in}
\begin{IEEEbiography}[{\includegraphics[width=1in,height=1.25in,clip,keepaspectratio]{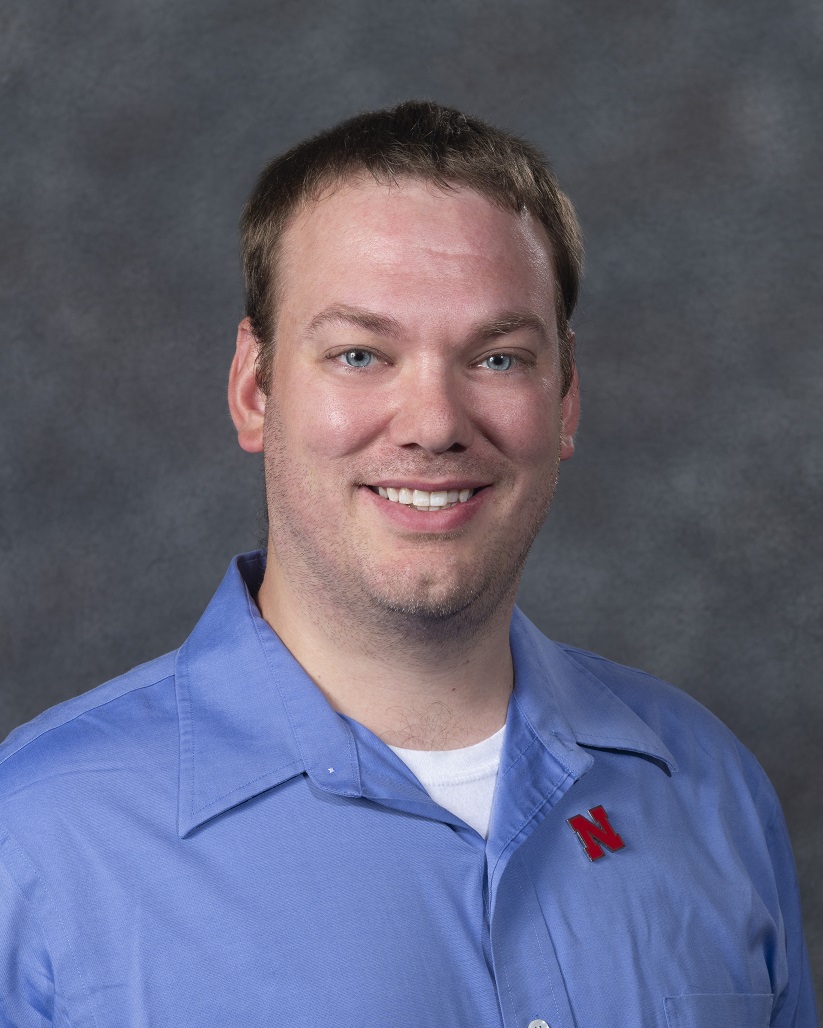}}]{Benjamin S. Riggan}{\space}(Member, IEEE) is currently an assistant professor with the Department of Electrical and Computer Engineering at the University of Nebraska-Lincoln (UNL).  Prior to joining the faculty at UNL, he was a research scientist at the U.S. Army Research Laboratory (ARL) in the Networked Sensing and Fusion branch.  His research interests are in the areas of computer vision, image and signal processing, and biometrics/forensics, especially related to domain adaptation, multi-modal analytics, and machine learning.  Dr. Riggan received the B.S. degree in computer engineering from North Carolina State University in 2009, and M.S. and Ph.D. degrees in electrical engineering from N.C. State University in 2011 and 2014, respectively.  After finishing his Ph.D., he was awarded a postdoctoral fellowship at ARL’s Image Processing Branch. He has received three best papers awards (IEEE WACV 2016, BTAS 2016, WACV 2018) and currently serves as Associate Editor for IEEE TAES.
\end{IEEEbiography}

\end{document}